# Deep Learning to Automate Parameter Extraction and Model Fitting of Two-Dimensional Transistors

Robert K. A. Bennett,[1] Jan-Lucas Uslu,[2,3] Harmon F. Gault,[1] Asir Intisar Khan,[1] Lauren Hoang,[1] Tara Peña,[1] Kathryn Neilson,[1] Young Suh Song,[1] Zhepeng Zhang,[3,4] Andrew J. Mannix,[3,4] Eric Pop[1,3,4,5,*]

[1]Department of Electrical Engineering, Stanford University, Stanford, CA 94305, USA
[2]Department of Physics, Stanford University, Stanford, CA 94305, USA
[3]Stanford Institute for Materials and Energy Sciences, SLAC National Accelerator Laboratory, Menlo Park, CA 94025, USA
[4]Department of Materials Science and Engineering, Stanford University, Stanford, CA 94305, USA
[5]Department of Applied Physics, Stanford University, Stanford, CA 94305, USA
*Contact: epop@stanford.edu

This document is the authors' version of a manuscript published in *Research*.

The published article can be accessed at https://spj.science.org/doi/epdf/10.34133/research.1103.

The full citation for the published article is:

Bennett, R.K.A., Uslu, J.-L., Gault, H.F., Khan, A.I., Hoang, L., Peña, T., Neilson, K., Song, Y.S., Zhang, Z., Mannix, A.J., & Pop, E. Deep Learning to Automate Parameter Extraction and Model Fitting of Two-Dimensional Transistors. *Research* **9**, doi:10.34133/research.1103 (2026).

## Abstract

We present a deep learning approach to extract physical parameters (e.g., mobility, Schottky contact barrier height, defect profiles) of two-dimensional (2D) transistors from electrical measurements, enabling automated parameter extraction and technology computer-aided design (TCAD) fitting. To facilitate this task, we implement a simple data augmentation and pretraining approach by training a secondary neural network to approximate a physics-based device simulator. This method enables high-quality fits after training the neural network on electrical data generated from physics-based simulations of ~500 devices, a factor >40× fewer than other recent efforts. Consequently, fitting can be achieved by training on physically rigorous TCAD models, including complex geometry, self-consistent transport, and electrostatic effects, and is not limited to computationally inexpensive compact models. We apply our approach to reverse-engineer key parameters from experimental monolayer $WS_2$ transistors, achieving a median coefficient of determination ($R^2$) = 0.99 when fitting measured electrical data. We also demonstrate that this approach generalizes and scales well by reverse-engineering electrical data on high-electron-mobility transistors while fitting 35 parameters simultaneously. To facilitate future research on deep learning approaches for inverse transistor design, we have published our code and sample data sets online.

## Introduction

Emerging semiconductors, such as two-dimensional (2D) transition metal dichalcogenides (TMDs),[1] amorphous oxides,[2] and carbon nanotubes,[3] show great promise to enable next-generation electronic devices. Field-effect transistors (FETs) based on these materials are frequently characterized by fitting their measured electrical characteristics to models that express current in terms of either material-level parameters (e.g., semiconductor mobility, Schottky contact barrier height) or device-level parameters (e.g., lumped contact resistances or parasitic capacitances). An advantage of this *model fitting* approach is that it simultaneously extracts multiple relevant quantities from a few measurements while enabling follow-up simulations to assess performance limitations or circuit-level analyses. However, fitting models to measured data tends to be an iterative process, even for experts in device simulations, making this powerful characterization tool both time-consuming and inaccessible to many researchers.[4]

To address these limitations, recent studies have developed neural networks that can predict model parameters from current *vs*. gate-to-source voltage and/or capacitance *vs*. gate-to-source voltage characteristics, achieving excellent fits after training these models on simulated data from 20,000 – 1,000,000 unique devices.[5-10] Acquiring such large training sets is easy for established technologies with computationally inexpensive models: for example, established compact models for gallium nitride[11] or silicon[12] transistors can compute hundreds or thousands of simulated electrical measurements per second. However, emerging FETs are frequently modeled using computationally expensive simulations[13-15] – often because the physics of such devices is more difficult to approximate, or because model development has lagged behind mature technologies – and can require minutes to hours[16,17] to generate a single current *vs*. voltage curve. For this reason, acquiring large data sets for training and fitting emerging device behavior is often intractable.

In this work, we develop a deep learning approach for FET parameter extraction that requires significantly less training data compared to other recent efforts. We reduce the amount of expensive training data by implementing a secondary surrogate network to approximate the electrical model, which we use to generate a large (and computationally inexpensive) augmented data set that we use to pretrain our primary network. Afterwards, we fine tune the primary network using data generated by the original physics-based model. This approach, from start to finish, requires generating electrical data from only ~500 devices (>40× fewer compared to previous works) to achieve excellent reverse-engineered fits to experimental data.

We verify our approach using both simulated and experimental test sets, obtaining median coefficients of determination ($R^2$) fits of 0.995 and 0.990, respectively, when fitting eight model parameters. We further demonstrate that our approach works for other types of FETs and scales well as the electrical models become more complex (i.e., have more fitting parameters). To facilitate future research, we have also published our data sets and machine learning code (implemented in Python using TensorFlow[18]) in a GitHub repository.[19]

## Results and discussion

**Overview of the deep learning approach:** In a standard electrical modeling task, the user inputs a set of model parameters, $Y_{\text{given}}$ (e.g., carrier mobility, Schottky contact barrier height, etc.), into a *forward model*, $\boldsymbol{M}_{\textbf{forward}}$, to compute a current $I_{\text{modeled}}$:

$$\boldsymbol{M}_{\mathbf{forward}}\left(Y_{\text{given}}\right) = I_{\text{modeled}} \tag{1}$$

Here, we aim to train a neural network for the inverse problem.[20] That is, given the measured current $I_{\text{measured}}$, we aim to train a neural network $\boldsymbol{N}_{\mathbf{inverse}}$ to predict the set of parameters $Y_{\text{pred}}$:

$$\boldsymbol{N}_{\mathbf{inverse}}(I_{\text{measured}}) = Y_{\text{pred}} \tag{2}$$

such that our original forward model reproduces the measured current:

$$I_{\text{measured}} \approx \boldsymbol{M}_{\mathbf{forward}}[\boldsymbol{N}_{\mathbf{inverse}}(I_{\text{measured}})] = I_{\text{modeled}} \tag{3}$$

Our approach for accomplishing this task is summarized in **Figure 1** and below:

1. Use a comprehensive physics-based model (e.g., Sentaurus Device[21]) to generate a training data set of input parameters and calculated electrical characteristics (**Figure 1A**).

2. Train an intermediate forward neural network to approximate the physics-based model (**Figure 1B**).

3. Generate a much larger augmented training set using the forward neural network and use this augmented data to pretrain the inverse neural network, calling the forward neural network repeatedly to guide training as a *tandem neural network* (**Figure 1C**).[22]

4. Finish training the inverse neural network using the original data set from step 1, once again calling the forward neural network repeatedly to guide training as a tandem neural network (**Figure 1D**).

5. Test the inverse neural network by inputting new electrical data, extracting the model parameters, and calling the physics-based model on the predicted model parameters to assess the accuracy of the extraction (**Figure 1E**).

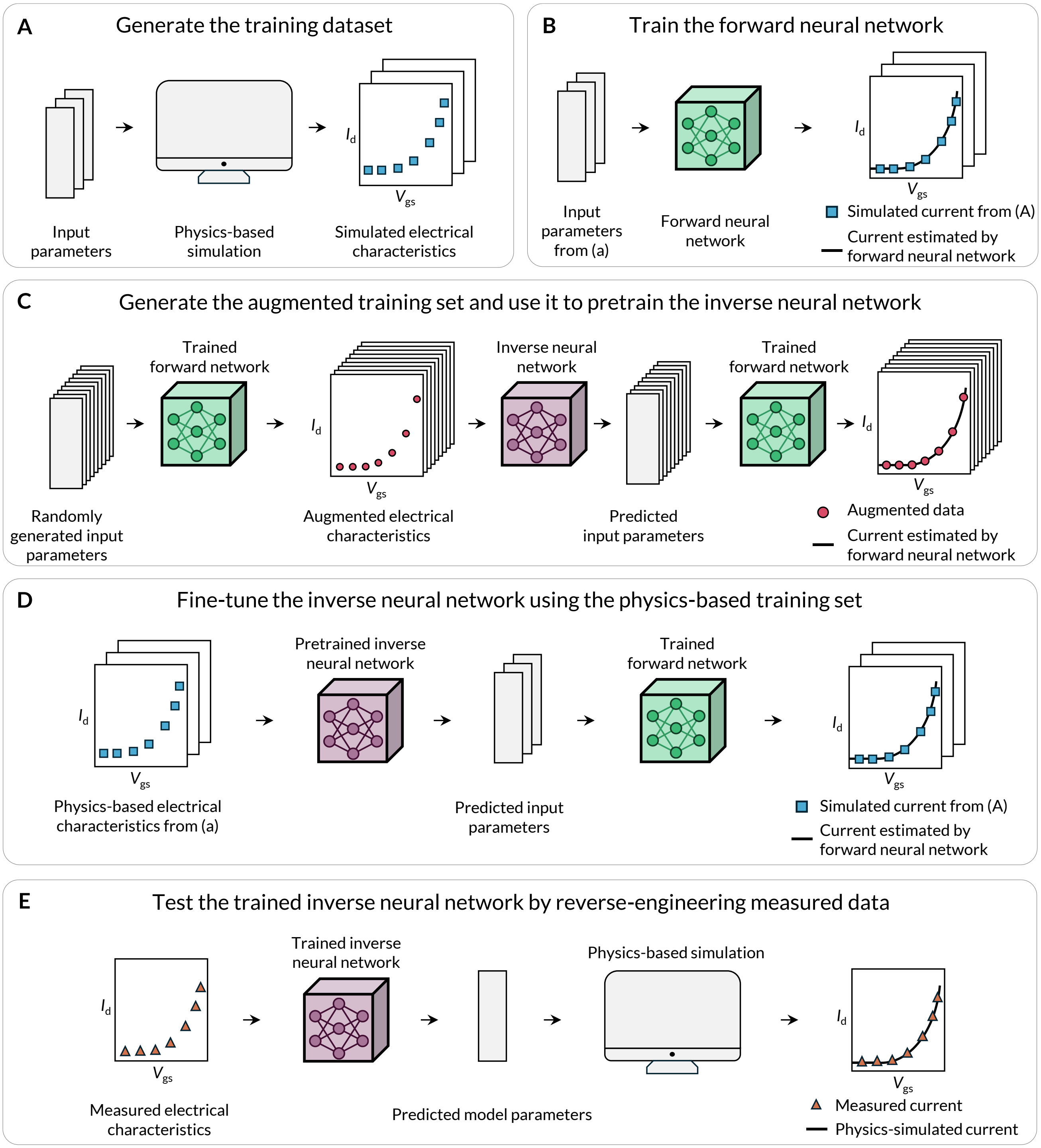

**Figure 1: Overview of our deep learning approach for reverse-engineering FET model parameters**. We begin by **(A)** generating a training set of model input parameters and output electrical characteristics from a physics-based device simulator (here, we use Sentaurus Device[21]). **(B)** We then train a forward neural network to reproduce the data from the physics-based simulator and **(C)** use this forward network to generate an augmented data set with which we pretrain the inverse neural network, then we finish training on the original physics-simulated electrical data **(D)**. In (**C**) and (**D**), we train the inverse neural network to predict model parameters that will reproduce the original input current; we estimate the current during training by calling the pretrained forward neural network from step (**B**) to do so (i.e., as a tandem neural network). **(E)** Next, we test the trained inverse neural network by providing it with experimental electrical data to be reverse-engineered, extracting the model parameters, and running new physics-based simulations using the predicted parameters. Finally, we plot the new simulated curves alongside the original measured data to assess the quality of the fit.

**<u>Generating the training set with a physics-based simulator</u>**: We generate our training data using Sentaurus Device,[21] an industry-standard technology computer-aided design (TCAD) program, as our physics-based model. This simulator can be any software that generates complete current *vs.* gate-to-source voltage data based on physical input parameters such as device geometry and material properties; other examples include Monte Carlo[23] or quantum transport simulators.[14,15]

For any fit to provide valid physical insights, regardless of how the fit is achieved – through conventional techniques or otherwise – the user must take care to ensure that the model used to generate the training data accurately describes their transistors. For instance, with enough fitting parameters, many models could enable a close fit between measured and modeled data, even if the physics of the model are wrong. However, for some applications, this could also be acceptable: for instance, empirical "phenomenological" models[24] and surrogate models[25] can be practical for circuit-level SPICE[26] simulations. In this work, we aim to provide an automated method to achieve a fit to a model *that the user specifies*, rather than to automate the process of selecting this model. For this fit to provide meaningful physical insight, it is therefore essential that the user select a physically meaningful model.

In our work, we use Sentaurus Device to generate $I_d$-$V_{gs}$ curves (where $I_d$ is the drain current and $V_{gs}$ is the gate-to-source voltage) for 2D FETs, using the device geometry shown in **Figure 2A**. This *back-gated* geometry is frequently used in experimental studies on 2D or other emerging semiconductors because it is easier to fabricate than top- or dual-gated devices; we use this structure here because we test our neural networks on readily available experimental $I_d$-$V_{gs}$ data from monolayer semiconductor transistors. We include additional Sentaurus Device simulation details in **Supplementary Section 1**. We discuss how this approach may be generalized to other types of transistors later in this work.

To build our training set, we initialize *n*-type FETs in Sentaurus Device TCAD[21] with the geometry shown in **Figure 2A**, choosing the value of each model parameter at random from the ranges listed in **Table 1**. Here, we intentionally use large ranges for each parameter to ensure the neural network can fit a wide range of device data, as the neural network cannot predict the values of parameters outside of these ranges. We emphasize that the eight parameters in **Table 1** are the *fitting* parameters considered in this work. Other semiconductor-level parameters (e.g., band gap, dielectric permittivity) and device-level parameters (e.g., channel length, oxide thickness) are treated as *fixed* model inputs for all simulations, as described in **Supplementary Section 1**.

With this approach, we then simulate sets of $I_d$-$V_{gs}$ characteristics for 25,000 devices from $V_{gs}$ = -6 to 50 V at $V_{ds}$ = 0.1 and 1.0 V, creating a large set of established $I_d$-$V_{gs}$ *vs.* model input parameter relationships with which we train our neural networks. We will later use subsets of this larger training set in bootstrap studies to determine the minimum acceptable training set size.

We note that if the distribution of any individual parameter is narrower than listed here, then the ranges of these parameters should be adapted accordingly. In general, reducing the allowable range of fitting parameters will make the fitting task easier; however, doing so requires one to be confident in the range of values a given parameter may take, which may not always be possible, especially for emerging semiconductor devices with unoptimized fabrication protocols.

It is also essential to note that the data used to train the neural networks must reflect the technology that the networks will be applied to. For instance, to build a neural network that will be used for the inverse design of another 2D semiconductor, the TCAD model used to build the training set should be updated to reflect the specifics of that semiconductor, and the parameters in **Table 1** should be adjusted to ensure they lie within their expected values for the experimental devices of interest.

**Table 1: Model parameters and the ranges across which they vary in the physics-based training set.** $N_{D0}$, $E_{D,mid}$, and $\sigma_D$ characterize donor-like defects, and $N_{A0}$ and $\sigma_A$ characterize acceptor-like band-tail states. The defect-like and acceptor-like profiles are shown in **Figures 2B,C**, respectively. We elaborate upon the physical meaning of these parameters and justify these selected ranges based on experimental and theoretical considerations in **Supplementary Section 2**.

| Parameter | Range | Unit |
|---|---|---|
| Mobility | 1 – 35 | $cm^2\ V^{-1}\ s^{-1}$ |
| Schottky contact barrier height | 10 – 510 | meV |
| Effective density of states $N_C$ | $2\times10^{11} – 9\times10^{12}$ | $cm^{-2}$ |
| Peak donor density $N_{D0}$ | $3\times10^{12} – 3\times10^{13}$ | $eV^{-1}\ cm^{-2}$ |
| Donor energy mid $E_{D,mid}$ | 20 – 200 | meV below $E_C$ |
| Donor energy width $\sigma_D$ | 20 – 200 | meV |
| Peak acceptor band tail density $N_{A0}$ | $6\times10^{11} – 3.7\times10^{13}$ | $eV^{-1}\ cm^{-2}$ |
| Acceptor band tail energy width $\sigma_A$ | 50 – 300 | meV |

**Model parameters for two-dimensional transistors**: The parameters we aim to fit include the electron mobility $\mu$, Schottky contact barrier height $\phi_B$, and effective density of states at the conduction band edge $N_C$. These parameters capture, respectively, the speed at which electrons can move through a semiconductor, the energy barrier associated with electron injection between the semiconductor and metal, and the number of available states for electrons in the semiconductor. The remaining parameters quantify defect profiles in 2D devices. Donor states (e.g., chalcogen vacancies; see **Supplementary Section 2**) follow a truncated Gaussian distribution (**Figure 2B**) with respect to energy $E$,

$$N_D(E) = N_{D0} \exp\left(-\frac{(E-E_{D,mid})^2}{2\sigma_D^2}\right) \text{ for } E < E_C \qquad N_D(E) = 0 \text{ otherwise} \qquad (4)$$

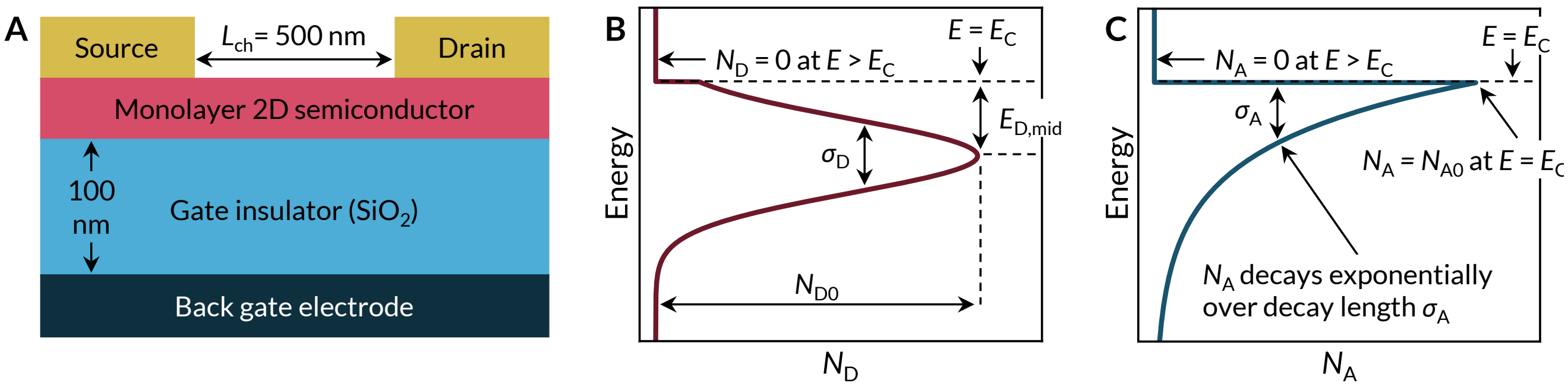

**Figure 2: Summary of transistors considered in this work**. (**A**) Diagram of our transistor geometry, showing a two-dimensional semiconductor integrated into back-gated field-effect transistor (not to scale). (**B**) The donor-like defect profile $N_D$ and (**C**) the acceptor-like band-tail $N_A$. These profiles are given **in Eqs. (4) and (5)**, respectively, and are parameterized in terms of the model parameters listed in **Table 1**.

where $N_{D0}$ is the peak donor density, $E_{D,mid}$ is the energy at which the Gaussian donor distribution peaks (referenced to the conduction band edge $E_C$), and $\sigma_D$ is the standard deviation (width) of the Gaussian curve.

We also include trap-like acceptor states (**Figure 2C**), which could arise from dispersionless band-tail states (see **Supplementary Section 2**). Their distribution decays exponentially from the conduction band edge $E_C$:

$$N_A(E) = N_{A0} \exp\left(-\left|\frac{E-E_C}{\sigma_A}\right|\right) \text{ for } E < E_C \qquad N_A(E) = 0 \text{ otherwise} \qquad (5)$$

where $N_{A0}$ is the peak acceptor density and $\sigma_A$ is the characteristic decay energy. We elaborate upon the physical meaning of each fitting parameter in **Supplementary Section 2**.

In this work, we use only synthetic (i.e., TCAD and compact model) data to train our neural networks. In principle, experimental data could also be incorporated into these training sets, although it may be difficult to obtain a sufficiently large labeled data set with purely experimental data, particularly for 2D semiconductors today. To incorporate experimental data when training a similar model, one could therefore use a "hybrid" data set, i.e., one with both experimental and TCAD data, while oversampling the experimental data.[27]

**Training the inverse neural network in tandem with a forward solver**: The inverse neural network architecture used in this work is shown in **Figure 3A**. This neural network accepts $I_d$-$V_{gs}$ data from physics-based simulations ($V_{ds}$ = 0.1 and 1.0 V, with 32 $V_{gs}$ values evenly spaced from -6 to +50 V); we also perform simple feature engineering by explicitly inputting $\partial I_d/\partial V_{gs}$, $\log_{10}(I_d)$, and $\partial \log_{10} I_d/\partial V_{gs}$ for each $V_{ds}$, yielding 8 vectors that we input into the neural network as a 32 × 8 matrix. We discuss and justify this feature engineering in **Supplementary Section 3**. Other recent works on machine learning-based parameter extraction have also included capacitance-voltage data as input features for silicon-based devices.[9] Although including such data would likely facilitate training and fitting, these capacitances are not consistently measured for 2D devices today, unlike $I_d$-$V_{gs}$ characteristics. To ensure our approach remains applicable to experimental 2D transistors without demanding additional measurements, we avoid relying upon these capacitance measurements here. Furthermore, while extremely low currents (e.g., <1 pA/μm) are readily simulated, they remain difficult to measure experimentally due to noise floors imposed by hardware. We therefore intentionally impose an artificial noise floor on all training data, as discussed in **Supplementary Section 4**, to avoid training our neural networks on these difficult-to-measure data.

The inverse neural network aims to output the set of Sentaurus model parameters associated with these $I_d$-$V_{gs}$ curves and derivatives. To accomplish this, we use the *tandem neural network* approach summarized in **Figure 3B**. Such tandem neural networks have found widespread use in many inverse design tasks,[22,28,29] including transistor parameter extraction.[7] In this approach, the inverse neural network is trained by feeding its output (i.e., the Sentaurus Device model parameters) into a pretrained forward neural network, which then outputs the estimated $I_d$-$V_{gs}$ characteristics corresponding to those model parameters. In other words, the forward neural network approximates the original physics-based TCAD solver or, equivalently, it approximates $\boldsymbol{M}_{\mathbf{forward}}$ in **Eq. (1)**. We describe all scaling, preprocessing, and training procedures in the **Materials and Methods** under **Neural Network Training Details** and in **Supplementary Section 4**. We describe the architecture of the forward neural network and characterize its performance in detail in **Supplementary Section 5**.

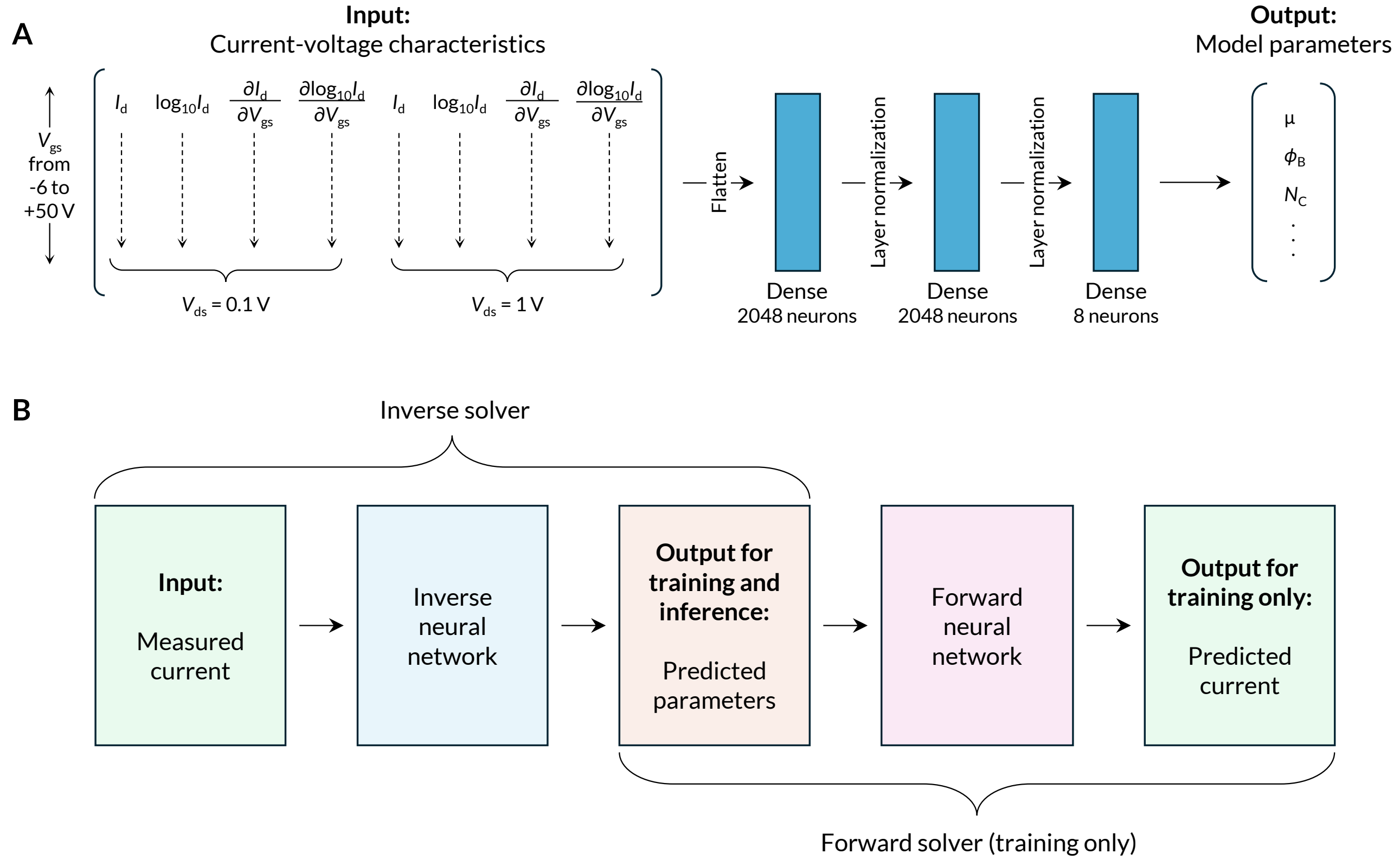

**Figure 3: Inverse neural network and tandem network training approach**. (**A**) Architecture of the inverse neural network. Each dense layer uses the rectified linear unit (ReLU) activation function, except the final layer, which uses the hyperbolic tangent (tanh) activation function. The network accepts current *vs*. gate-to-source voltage ($I_d$-$V_{gs}$) measurements in linear and logarithmic space, along with their derivatives with respect to $V_{gs}$, and outputs the set of model parameters that a physics-based solver can use to reproduce this input data. (**B**) The tandem approach used for training the inverse neural network. During training, the output of the inverse neural network is fed into a pretrained forward neural network, which allows us to include errors of predicted current in the loss function. We use this tandem approach only while training; during inference, the network outputs the predicted parameters directly. Details of the training procedures and descriptions of the forward neural network are given in the **Materials and Methods** and in **Supplementary Sections 4 and 5**.

A tandem approach is necessary here because $I_d$-$V_{gs}$ characteristics are generally not unique functions of the model parameters.[7] In other words, there are often multiple sets of input parameters that can allow the physics-based simulator to yield identical or near-identical $I_d$ profiles, making it difficult to train the inverse neural network based on errors in parameters alone. Instead, training the inverse neural network using a tandem approach allows us to incorporate errors in both parameters and predicted current into the loss function,

$$\mathcal{L} = \sqrt{\left(\frac{1}{N_{\text{devices}}}\right)\left(\frac{1}{N_{\text{params}}}\right)\sum_{i}^{N_{\text{devices}}}\sum_{\ell}^{N_{\text{params}}}\left(Y_{\text{actual}}^{(i,\ell)} - Y_{\text{pred}}^{(i,\ell)}\right)^2 \cdot \frac{1}{N_{\text{devices}}}\sum_{i}^{N_{\text{devices}}} E_{I_{\text{d}}}^{(i)}} \quad (6)$$

where $N_{\text{devices}}$ is the number of devices, $N_{\text{params}}$ is the number of fitting parameters in the parameter set $Y$, the subscripts "actual" and "pred" denote baseline truths and predicted quantities, and the superscripts ($i$) and ($\ell$) denote the $i^{\text{th}}$ device and $\ell^{\text{th}}$ fitting parameter, respectively. $E_{I_{\text{d}}}^{(i)}$ is the error in $I_d$ between the actual and predicted currents and their derivatives, given by **Supplementary Eq. (S1)** in **Supplementary Section 4**. During training, we estimate the predicted current $I_{\text{pred}}^{(i)}$ by calling the forward network on the predicted parameter set, i.e., $I_{\text{pred}}^{(i)} = \boldsymbol{M}_{\text{forward}}\left(Y_{\text{pred}}^{(i)}\right)$.

We emphasize that we use this tandem approach only for *training* the inverse neural network, not for inference.

**Pretraining with an augmented data set and fine-tuning with physics-based data**: We take advantage of the forward neural network to build an augmented data set that we use to pretrain our inverse network. Here, we generate many sets of randomly chosen input parameters using the same parameters and ranges from **Table 1**, then use the forward network to estimate the corresponding $I_d$-$V_{gs}$ curves. These augmented data are essentially free compared to our original physics-based simulated data: it takes ~1 minute to generate a single $I_d$-$V_{gs}$ curve with Sentaurus (for other device simulators, this time can vary greatly, depending on the specific device and models used), whereas we can generate 100,000 $I_d$-$V_{gs}$ curves with the forward neural network in less than 30 seconds. After pretraining our inverse neural network on this augmented data set, we use the physics-based simulation training set to fine-tune the model. We detail full training procedures in **Supplementary Section 4**.

After training, we test the performance of the inverse neural network by using it to extract estimated model parameters from a Sentaurus Device-simulated test of 1,000 devices (generated with the same procedure as training and development sets) that were unseen by the neural network during training and network optimization. Afterwards, we take these extracted parameters and estimate their corresponding $I_d$-$V_{gs}$ curves using a well-trained forward neural network that was trained on all 25,000 devices in our original physics-based training set. We show in **Supplementary Section 5** that this forward neural network reproduces our original Sentaurus simulations with exceptional accuracy; thus, this approach allows us to estimate the accuracy of the inverse neural network without having to re-run Sentaurus Device simulations, which can become computationally expensive when repeatedly evaluating the performance of the inverse neural network on multiple training sizes. Finally, we calculate $R^2$ between the original and newly obtained $I_d$-$V_{gs}$ characteristics to quantify how accurately these estimated parameters reproduce the original $I_d$-$V_{gs}$ curves (higher $R^2$ is better; $R^2 = 1$ is a perfect fit; details of the $R^2$ calculation are in **Supplementary Section 4**).

Our original physics-based training set contains $I_d$-$V_{gs}$ curves from 25,000 unique devices; however, we wish to develop a machine learning approach that relies upon as few simulations as possible, since they are computationally expensive. Thus, we apply the bootstrap approach described in **Supplementary Section 4** to draw smaller training sets to quantify how the training set size impacts the performance of the neural network. Note that when we refer to the training set size, we refer to the number of device configurations used for physics-based simulations in both the training and development sets combined and that for each bootstrapped training procedure, we use the same subset of data to train both the forward and inverse networks. For example, if we specify that the training set size is 500, we (i) use 500 sets of physics-based simulations to train the forward neural network; (ii) use that forward neural network to generate a large augmented data set (100,000 sets of $I_d$-$V_{gs}$ curves) with which we train the inverse neural network; and then (iii) fine-tune the inverse neural network on the original 500 physics-based simulations.

We plot the 5th quantile (worst 5%) $R^2$ for various training set sizes, with and without pre-training, in **Figure 4A**, and show sample histograms for training sets consisting of 500 and 1,000 devices in **Figure 4B**. Here, we find that pretraining significantly improves the performance of the inverse neural network; in terms of $R^2$ fit, implementing this pretraining procedure appears to offer an accuracy boost roughly equivalent to doubling the size of the original physics-based training set. We show sample fits corresponding to the 5th quantile (worst 5%), 10th quantile (worst 10%), and median $R^2$ in **Figures 4C-E** for an inverse neural network with pretraining (500 devices in the physics-based training set), confirming a good match between the original simulated $I_d$-$V_{gs}$ data and that obtained with the reverse-engineered parameters.

Our approach sharply reduces the number of physics-based (Sentaurus Device TCAD) simulations needed to achieve excellent fits across the majority of the test set compared to previous studies. These past efforts have used training sets consisting of 20,000 – 1,000,000 devices to achieve high-quality fits for a similar reverse-engineering task, whereas we demonstrate excellent fits using training sets of ~500 devices, reducing the required number of training devices by ~40×. For this training set size, the full training procedure takes less than five minutes (including training the forward network, generating the augmented data set, and then pretraining and fine tuning the inverse neural network) on an NVIDIA RTX 4080 graphics processing unit, which is much less time compared to that required to generate the physics-based data.

The pretraining approach we implement is especially beneficial when fitting data to computationally expensive TCAD models, since we reduce the number of time-consuming simulations required to build the training set. This pre-training approach is less useful when fitting computationally inexpensive models (e.g., compact models implemented in Verilog-A), where large data sets are easily and cheaply obtained. However, even when simulated $I_d$-$V_{gs}$ data are inexpensive, we emphasize that the surrogate model still facilitates training by providing a GPU-compatible means to evaluate the $E_{I_d}^{(i)}$ term in the loss function [**Eq. (6)**]. Therefore, training a surrogate model (as in **Figure 1B**) remains advisable for this inverse design approach, even when the target model is computationally inexpensive.

This reduction in training set size makes it feasible to train the neural network to train on data generated by computationally-intensive models: for example, assuming parallelization to 32 cores, it would take ~30 minutes to generate a 500 device training set if using a TCAD solver like Sentaurus with ~two minutes per $I_d$-$V_{gs}$ curve, or ~1.25 days if using a quantum-transport solver (e.g., the non-equilibrium Green's function method) with ~one hour per $I_d$-$V_{gs}$ curve. It would take ~1 day or ~1 month (respectively) to generate these training sets if using a fitting approach that requires training on 20,000 devices; thus, the approach presented here reduces a day-long generation task to something that can be accomplished in under an hour, or an (often) unfeasibly long task to something that can be accomplished in a day.

Next, we investigate the accuracy of the extracted parameters, plotting the actual *vs*. predicted values for mobility, Schottky contact barrier height, and peak donor density in **Figures 4F-H**. We show similar plots for all model parameters in **Supplementary Section 6**. Here, we find that the mobility and Schottky barrier height can both be extracted with relatively low median absolute errors of 0.88 $cm^2 V^{-1} s^{-1}$ and 18.4 meV, respectively, representing 2.6% and 3.7% of their nominal ranges. The standard deviation of the mobility error is also quite small at 1.9 $cm^2 V^{-1} s^{-1}$ (5.6% of its nominal range), whereas the standard deviation for the Schottky barrier height (47.1 meV) is 9.4% of its nominal range. However, from **Figure 4G**, most of these errors occur when the actual Schottky barrier height is comparable to the thermal energy $k_B T$ (where $k_B$ is the Boltzmann constant and $T$ = 300 K is the absolute temperature), which makes sense physically: once the barrier is "small enough," small deviations will not meaningfully affect the contact resistance or the resultant $I_d$.

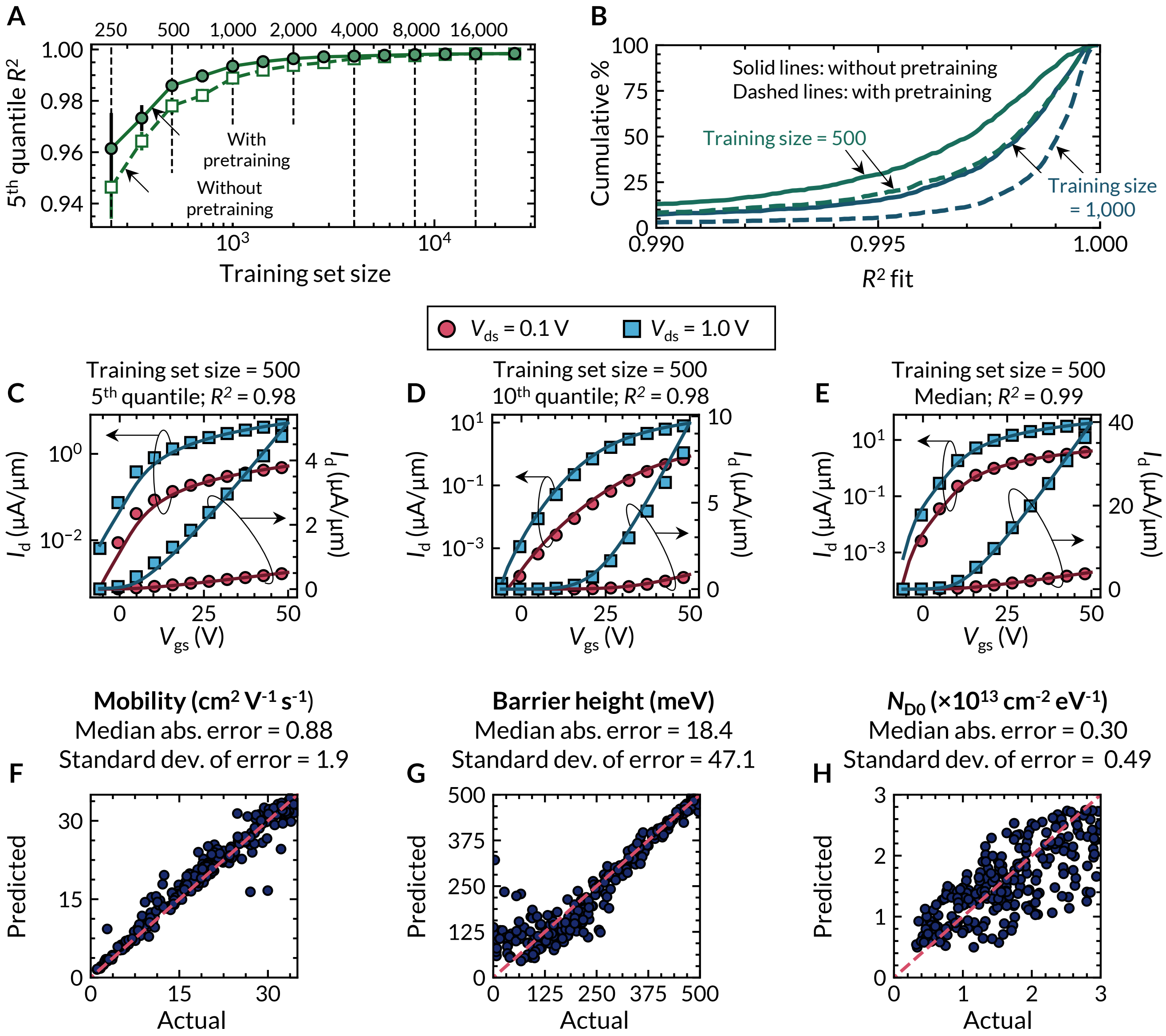

**Figure 4: Performance of the inverse network on the simulated test set.** (**A**) The 5th quantile (i.e., worst 5%) $R^2$ fits attained by the inverse neural network as a function of training size, with and without pretraining, and (**B**) sample cumulative $R^2$ fits achieved by the inverse neural network, without (solid lines) and with (dashed lines) pretraining. Error bars in (A) show standard deviations across five bootstrapped runs; error bars are often small and are not always visible. Sample current *vs*. gate-to-source voltage fits after training the inverse neural network on 500 devices, showing the reverse-engineered fits corresponding to (**C**) the 5th quantile, (**D**) 10th quantile, and (**E**) median $R^2$ (symbols: original physics-based simulations; lines: estimated $I_d$-$V_{gs}$ curves obtained using reverse-engineered parameters). The data are shown on both logarithmic (left) and linear axes (right), at $V_{ds}$ = 0.1 V (red) and 1.0 V (blue). Actual *vs*. predicted values for (**F**) mobility, (**G**) Schottky contact barrier height, and (**H**) peak donor density. The pink dashed lines are the lines of zero error, i.e., where predicted and actual values match perfectly.

We find larger errors in the other model parameters – for example, as shown in **Figure 4H**, the median absolute error for the peak donor density $N_{D0}$ (3.0 × $10^{12}$ $cm^{-2}$ $eV^{-1}$) represents 11.1% of its nominal range, and the relatively large standard deviation (4.9 × $10^{12}$ $cm^{-2}$ $eV^{-1}$) represents 18.1% of this range. We speculate that we see larger errors in these parameters because they have some mutual redundancy: for example, if the peak acceptor density is underestimated, then the overall acceptor profile could still be approximated by increasing its energy width to compensate, as we demonstrate in **Supplementary Section 6**. In comparison, the mobility and Schottky contact barrier height both influence the $I_d$-$V_{gs}$ curves uniquely compared to other parameters. In other words, if these parameters are estimated incorrectly, the overall $I_d$-$V_{gs}$ curves cannot be matched by adjusting other model parameters to compensate, leading to lower error in their extractions.

We note that for some applications, like circuit-level modeling, large errors in individual parameters may be acceptable as long as the fitted model can accurately reproduce experimental $I_d$-$V_{gs}$ (and/or $I_d$-$V_{ds}$) data. Alternatively, for applications where strong physical insights are needed – such as monitoring the quality of a growth by estimating the defect density in the semiconductor – the accuracy of individual parameters for a specific device may be essential. We discuss strategies for estimating the uncertainty for these individual parameters in **Supplementary Section 6**.

**Experimental validation using measured $I_d$-$V_{gs}$ data from monolayer $WS_2$ transistors**: We validate our deep learning approach experimentally by using a test set collected from 52 monolayer $WS_2$ transistors that were fabricated across 3 separate chips.[30] All measured working devices were included, except for one outlier that could not achieve an on-state current of at least 1 μA/μm at $V_{gs}$ = 50 V and $V_{ds}$ = 1 V, leaving a total of 51 devices in the test set. These devices follow the geometry shown in **Figure 2A** with a channel length of 500 nm, 100 nm $SiO_2$ back-gate insulators, and nickel contacts (full experimental details are described in the **Materials and Methods** under **Experimental Details for Monolayer $WS_2$ Transistors**). We choose this channel length because it offers an intermediate regime where both the channel and contacts will influence $I_d$, allowing us to characterize both the channel and contacts. To match our physics-based training sets, we perform $I_d$-$V_{gs}$ sweeps at $V_{ds}$ = 0.1 and 1 V (all curves at $V_{ds}$ = 1 V are plotted in **Figure 5A**) and then feed the measured data into our inverse network after training it on physics-based simulations from 500 devices. We then extract the 8 model parameters listed in **Table 1** and perform new Sentaurus TCAD simulations using these predicted parameters.

We plot the cumulative $R^2$ between the TCAD-simulated $I_d$-$V_{gs}$ curves (simulated using the predicted input parameters) and original experimental data in **Figure 5B,** finding that the 25$^{th}$ quantile (worst 25%), median, and 75$^{th}$ quantile (best 25%) $R^2$ fits for the test sets are 0.984 (95% confidence interval: 0.979-0.987), 0.990 (95% confidence interval: 0.986-0.993), and 0.994 (95% confidence interval: 0.992-0.996), respectively, where confidence intervals are estimated by the empirical bootstrap method.[31] We plot the corresponding $I_d$-$V_{gs}$ fits in **Figures 5C-E**, along with fits for all 51 experimental devices in **Supplementary Section 7**.

A key advantage of this deep learning approach is that once the neural networks are trained, we can rapidly characterize many devices simultaneously to study variation across one or more chips. For example, we plot the cumulative histogram of the maximum measured current at $V_{gs}$ = 50 V and $V_{ds}$ = 1 V, $I_d^{(max)}$, in **Figure 5F**, as well as the extracted electron mobility and Schottky contact barrier heights in **Figures 5G,H**. We find that the extracted mobility and barrier heights show significant variation, agreeing well (qualitatively) with the experimental variation in **Figures 5A,F**. Overall, we anticipate that this approach could be used to facilitate characterization and analysis of large distributions of devices, so long as the ranges of variables in the training set are sufficiently broad to encapsulate the variation of these relevant parameters experimentally.

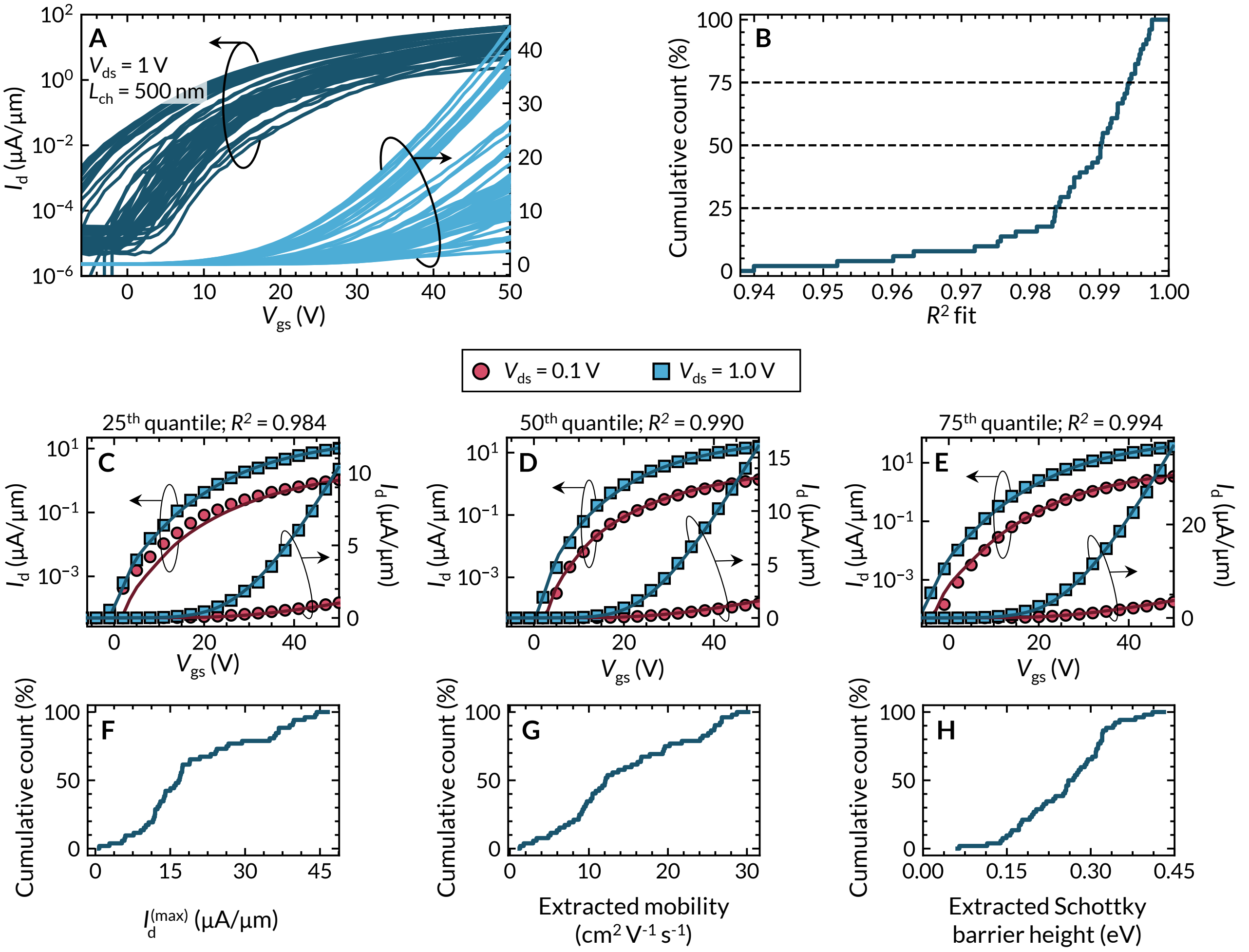

**Figure 5: Performance of the inverse network on experimentally measured test devices.** (**A**) All 51 measured $I_d$-$V_{gs}$ curves for monolayer $WS_2$ transistors ($V_{ds}$ = 1 V) in our experimental test set, shown both on logarithmic (left) and linear (right) *y*-axes. (**B**) Cumulative histogram showing the $R^2$ fits between the original experimental data and devices reverse-engineered by an inverse neural network trained on 500 sets of physics-based simulations. Sample $I_d$-$V_{gs}$ fits for (**C**) the 25th quantile, (**D**) median, and (**E**) 75th quantile $R^2$, showing good agreement between the experimental data (symbols) and our reverse-engineered fits (lines). Data are shown on both logarithmic (left) and linear axes (right), at $V_{ds}$ = 0.1 V (red) and 1.0 V (blue). Cumulative histograms for (**F**) the experimental $I_d^{(max)}$, (**G**) the extracted mobility, and (**H**) the extracted Schottky barrier height.

**Figures 5G,H** also address questions about the distribution of electron mobility and contact resistance one can expect for a particular $WS_2$ synthesis and fabrication approach. For example, the mobility includes information about the electron effective mass and scattering mechanisms with lattice vibrations and defects, both in the $WS_2$ film and its substrate. In other words, **Figure 5G** could be used to obtain information about the defect distributions in such $WS_2$ films. Similarly, **Figure 5H** can be used to understand how the deposition of the metal (nickel) affects the transistor contacts. We caution that the distributions shown here are for unoptimized

academic devices. One must increase the mobility, decrease the Schottky barrier height, and tighten these distributions before these transistors can become competitive in the semiconductor industry.

We emphasize that in model fitting, a close match between the original data and the model fit (i.e., a fit that yields $R^2 \approx 1$) does *not* guarantee that the extracted parameters are necessarily accurate. For example, a high-quality fit could also be achieved by the model overfitting the data. This is a well-known limitation of model fitting that must be considered when interpreting the results of any parameter extraction (whether the fit is achieved manually, through conventional techniques, or using the deep learning method in this work). To partially address issues arising from overfitting, one could also incorporate temperature-dependent data, incorporate hysteresis explicitly, add additional $I_d$-$V_{gs}$ curves (at additional fixed $V_{ds}$ values), and/or add $I_d$-$V_{ds}$ curves (at multiple fixed $V_{gs}$ points), and enable electrical models computed self-consistently with device heating effects. These could reduce the number of parameter sets that yield near-identical fits across the entire range of data.

**Model generalization and fitting many physical parameters simultaneously**: Until now, we have confined our discussion to 2D semiconductor transistors simulated by TCAD using 8 fitting parameters. However, model fitting is routinely applied to other types of devices, often using highly dimensionalized models with many parameters.[11,12] In **Supplementary Section 8**, we study how our approach scales with the complexity of the $I_d$-$V_{gs}$ model by applying the same methodology to train a neural network to predict the input parameters of GaN high-electron-mobility transistors (HEMTs) simulated with the industry-standard ASM-HEMT compact model.[11] We find that training becomes more difficult when fitting more physical transistor parameters simultaneously; this result is an expected consequence of the *curse of dimensionality*[32] encountered when fitting high-dimensional data, as the size of the parameter space grows exponentially as additional parameters are added (e.g., a space with $N$ dimensions, each sampled along $k$ values, has size $k^N$). Nevertheless, we demonstrate that we can still achieve good fits when fitting 15 or 25 or 35 parameters with a training set containing data from ~1,000 or 4,000 or 16,000 unique devices, respectively.

These scaling trends indicate that the method proposed in this work could be applied to fit transistor data even to complicated models with many unknown parameters, as long as generating the appropriately sized training set remains computationally tractable. This consideration varies tremendously from model to model: for example, generating such large training sets could be trivial for compact models implemented in Verilog-A, including the ASM-HEMT model[11] considered above, which can be used to generate thousands of $I_d$-$V_{gs}$ curves per second. It may be significantly harder to generate such large data sets with rigorous physics-based models, especially in the face of hardware limitations and/or limited licenses, which may restrict parallelizability.

Once a neural network is fully trained, it can be used repeatedly to fit many devices with minimal computation cost, as we do with experimental devices in **Figure 5**. In other words, the process of acquiring training data is a large "up front" cost; however, once the model is fully trained, model inference (i.e., applying the inverse neural network to extract device parameters from $I_d$-$V_{gs}$ curves), takes only a small fraction of one second. Therefore, this deep learning approach is especially appealing when a model fitting procedure must be repeated multiple times, e.g., to monitor the performance of a device over time, to quantify the performance of devices on a production line, or to fit many devices in a large experimental data set simultaneously. Alternatively, if only a single device needs to be fit to a TCAD or compact model, then another fitting approach, e.g., manual fitting or derivative-free optimization,[4] could incur a lower total computational cost compared to the deep learning approach studied here.

## Conclusion

In conclusion, we have developed a deep learning approach to automate parameter extraction (TCAD fitting) for transistors based on their current *vs.* gate-to-source voltage characteristics. We significantly reduce the number of devices required in our training set compared to those demonstrated in prior works, and we demonstrate excellent fits on experimental 2D transistors using models trained on physics-based simulations from 500 devices. This approach also generalizes well to other types of transistors and models (e.g., GaN high-electron-mobility transistors with up to 35 unknown parameters in the ASM-HEMT model), as long as acquiring larger training sets remains computationally feasible. Thus, we anticipate that the results of this study will facilitate fitting experimental device measurements to both physics-based and compact models alike. Furthermore, machine learning models have recently been shown to be efficient for modeling emerging technologies, including static random access memory (SRAM),[33] three dimensional NAND memory,[34] and 2D terahertz transistors.[35] Therefore, the inverse design approach we present here could facilitate the design and analysis of these and other devices.

In this work, we have not investigated how noisy or partially corrupted measurements affect the accuracy of this method. To make this fitting method more robust against noise and corruption, these factors could be incorporated into the training data, e.g., by imposing artificial noise or by artificially corrupting data within the data sets used to train the neural networks. Further, engineering the loss function, e.g., as implemented in a previous work,[4] could offer another promising method of making this machine learning approach more robust.

A disadvantage of this machine learning framework (and similar approaches put forward in previous works[5-10]) is that fitting devices with different geometries (e.g., varying channel length or oxide thickness, or changing the gating configuration between bottom-, top-, or dual-gated devices) requires that training sets be regenerated and neural networks be retrained. To address this issue, one could explicitly include channel length and other relevant geometry parameters as flexible inputs when training the forward and inverse neural networks. Alternatively, neural networks could be adapted across geometries via transfer learning,[36] i.e., using a tandem network originally pretrained on one geometry, and then using less fine-tuning data to adapt it to a different device geometry. Both approaches are promising research directions for future work that would boost the robustness and practicality of the deep learning method implemented here.

To facilitate these future research directions, we have uploaded our data sets and Python code to an online GitHub repository.[19]

## Materials and methods

**Neural network training details**: We use the Adam Optimizer,[37] as implemented in TensorFlow (Keras),[18] for all training. We train using early stopping, saving only the best performing model, as measured by the validation loss on the development set. During training, we use learning rate annealing. Here, the annealing rate refers to the factor that we multiply the learning rate by at the beginning of each annealing step.

When training the forward neural network, we anneal across 4 steps with an annealing rate of 0.35, using an initial learning rate of $10^{-3}$ and an early stopping patience of 50. We use a minibatch size of 128 when training the forward network. When pretraining the inverse neural network, we anneal across two steps, each with no more than 50 epochs, using an initial learning rate of $2.5\times10^{-4}$, an annealing rate of 0.6, and an early stopping patience of 5. After pretraining, we fine-tune the inverse neural network on the original training set used to train the forward network. In this stage, we anneal the inverse neural network across 3 steps with an annealing rate of 0.9. We use an initial learning rate of $2\times10^{-5}$ and an early stopping patience of 40. When we train the inverse neural network without pretraining, as in **Figures 4A,B** in the main text, we use an initial learning rate of $10^{-3}$, an annealing rate of 0.8, 10 annealing steps, and a patience of 40 for early stopping. We always use a minibatch size of 256 when training the inverse network.

The pretraining data set uses 100,000 devices when training the inverse neural network to fit data from two-dimensional transistors (e.g., **Figures 4 and 5**). When training to fit data from HEMTs in **Supplementary Section 8**, the size of the pretraining data set is the smaller of (i) 100× the number of compact model simulations or (ii) 1,000,000. Additional computational details, including details for preprocessing and scaling, are described in **Supplementary Section 4**.

**Experimental details for monolayer $WS_2$ transistors**: Monolayer $WS_2$ was grown by hybrid metal-organic chemical vapor deposition and was wet-transferred onto 100 nm $SiO_2/p^{++}$ Si, which served as the back-gate. Polystyrene was spin-coated on top of the $WS_2$ and then transferred in deionized water onto the target substrate ($SiO_2$/Si). Polystyrene was then removed using toluene. Electron-beam lithography was employed for each lithography step. First, large probing pads (Ti/Pt 2/20 nm) were patterned and deposited by electron beam evaporation via lift-off. Channel definition was done using xenon difluoride etching and the contacts were patterned for liftoff. Ni/Au contacts were evaporated by electron beam at ~$10^{-8}$ Torr. The channel and contact lengths of all devices were 500 nm and 1.5 µm, respectively. The devices were measured in a Janis ST-100 probe station at ~$10^{-4}$ Torr using a Keithley 4200 semiconductor parameter analyzer. These devices and measurements were described in a previous work.[30]

Experimental devices were fabricated across three separate chips. We include all measured working devices in our test set, except for one device that could not achieve an on-state current >1 µA/µm at $V_{ds}$ = 1 V, leaving a total of 51 devices in our test set. Experimentally, we measure both forward and backward sweeps for each device using a $V_{gs}$ range of ±60 V; in this work, we use only the forward sweeps and restrict $V_{gs}$ from -6 to +50 V for the reverse-engineering process described in the main text.

## Funding

R.K.A.B. acknowledges support from the Stanford Graduate Fellowship (SGF) and NSERC PGS-D programs. J.U. acknowledges support from the US Department of Energy, Office of Science, Basic Energy Sciences, Materials Sciences and Engineering Division, under contract DE-AC02-76SF00515. H.F.G. acknowledges support from the Stanford Electrical Engineering Research Experience for Undergraduates (REU) program, the Stanford Undergraduate Research and Independent Projects Program in the form of a Small Grant, and the SUPREME Undergraduate Microelectronics Fellowship program. T.P. acknowledges support from the NSF Ascend Post-doctoral Fellowship. T.P. and E.P. also acknowledge Intel Corporation. K.N. acknowledges support from the Stanford Graduate Fellowship (SGF) and the National Science Foundation Graduate Research Fellowship Program (NSF-GRFP). Device fabrication was performed at nano@stanford (RRID:SCR_026695). The authors also acknowledge partial support from the Stanford SystemX Alliance and from SUPREME, one of seven centers in JUMP 2.0, a Semiconductor Research Corporation (SRC) program sponsored by DARPA.

## Author contributions

R.K.A.B. conceived the project idea. R.K.A.B. and Y.S. configured the Sentaurus TCAD model. R.K.A.B., J.U., H.F.G., developed and verified the machine learning models. L.H., T.P., Z.Z., and K.N. fabricated monolayer semiconductor devices. L.H., T.P., and K.N. performed electrical measurements. R.K.A.B., H.G., A.K., A.J.M., and E.P. analyzed and interpreted the results. E.P. and A.J.M. supervised the research. R.K.A.B. and E.P. wrote the manuscript with input from all authors. All authors read and approved the final version of the manuscript.

## Competing interests

The authors declare no conflicts of interest.

## Data availability

Sample training and test data used in this study are available in an online GitHub repository.[19] All other data that support the findings of this study are available from the corresponding author upon request.

## Code availability

Code that supports the findings of this study is available in an online GitHub repository.[19]

# Supplementary Information

# Deep Learning to Automate Parameter Extraction and Model Fitting of Two-Dimensional Transistors

Robert K. A. Bennett,[1] Jan-Lucas Uslu,[2,3] Harmon F. Gault,[1] Asir Intisar Khan,[1] Lauren Hoang,[1] Tara Peña,[1] Kathryn Neilson,[1] Young Suh Song,[1] Zhepeng Zhang,[3,4] Andrew J. Mannix,[3,4] Eric Pop[1,3,4,5,*]

[1]Department of Electrical Engineering, Stanford University, Stanford, CA 94305, USA
[2]Department of Physics, Stanford University, Stanford, CA 94305, USA
[3]Stanford Institute for Materials and Energy Sciences, SLAC National Accelerator Laboratory, Menlo Park, CA 94025, USA
[4]Department of Materials Science and Engineering, Stanford University, Stanford, CA 94305, USA
[5]Department of Applied Physics, Stanford University, Stanford, CA 94305, USA
*Contact: epop@stanford.edu

## Supplementary Section 1: Details for Sentaurus simulations of two-dimensional transistors

We use Sentaurus Device,[1] an industry-standard technology computer-aided design (TCAD) simulator, for the physics-based simulations used in this work. We perform simulations in a two-dimensional domain using the cross-section shown in **Figure 2A** from the main text, where the width dimension of the device (into the page, perpendicular to the channel) is treated as periodic. Here, the default drift-diffusion model for transport is solved self-consistently with electrostatics (Poisson's equation).

We treat the two-dimensional (2D) semiconductor/metal contacts as Schottky contacts using Sentaurus Device's Schottky barrier model, and we enable tunneling through the barrier using Sentaurus Device's Schrodinger model on a nonlocal mesh. Although Fermi level pinning is known to affect the barrier height of 2D semiconductor/metal contacts,[2] Sentaurus Device does not presently have an established model for this pinning. To achieve a specific nominal barrier height, we therefore use the 2D semiconductor's electron affinity as a free parameter, with the resultant barrier height given by $\phi_{B0} = \chi_s - \phi_m$, where $\chi_s$ is the 2D semiconductor's electron affinity and $\phi_m$ = 5.01 eV is the work function of nickel. Gate-voltage-dependent contact resistance, an important experimental consideration,[3] is incorporated in our Sentaurus Device simulations automatically, as the electric field from the back-gate electrode electrostatically influences the 2D semiconductor underneath the metal.

We assign the monolayer 2D semiconductor an electronic band gap of 2.2 eV (larger than the optical band gap dictated by the exciton binding energy),[4,5] a channel thickness of 0.62 nm,[6] and in-plane and out-of-plane relative permittivities of 14 and 7,[7] respectively. We approximate the density of states using Sentaurus Device's effective mass model (as implemented by setting Formula=2).[1] We treat donor-like states (often attributed to chalcogen defects in a 2D semiconductor like $WS_2$) as states located at the 2D semiconductor/oxide interface, and we assume band-tail acceptor-like states are uniformly distributed across the 2D channel thickness.

## Supplementary Section 2: Descriptions of model parameters for two-dimensional transistors

In **Table 1** of the main text, we list our fitting parameters alongside their lower and upper bounds. Here, we briefly explain the physical significance of each parameter and justify these bounds.

**Mobility**: A semiconductor's electron mobility is the proportionality constant between the electron speed and the applied electric field, assuming drift-diffusion transport (which is valid for the relatively long 500 nm channel lengths considered in this work). Typical academic 2D transistors often yield a mobility of the order ~10 $cm^2$ $V^{-1}$ $s^{-1}$; here, we use a range of 1 to 35 $cm^2$ $V^{-1}$ $s^{-1}$ to encapsulate this typical value. We note that electron mobilities above 35 $cm^2$ $V^{-1}$ $s^{-1}$ are certainly possible in optimized devices with fewer defects; to fit such devices, the training set should be expanded to incorporate these higher mobilities.

The mobility considered in this work implicitly lumps various sources of mobility degradation (e.g., grain boundary scattering, remote phonon or charged impurity scattering) into one unified mobility. With appropriate mobility models, these factors could be included within the set of fitting parameters in **Table 1** to study these mechanisms further in emerging semiconductor devices.

**Schottky barrier height**: The Schottky barrier height is the energy barrier which a carrier must overcome to be injected from the metal contact to the semiconductor channel or vice versa. The Schottky barrier height is a function of its electrostatic environment and is hence not an intrinsic material or device parameter.[8] When we report Schottky barrier heights in this work, we specifically refer to the *nominal* Schottky barrier height $\phi_{B0} = \chi_s - \phi_m$, where $\chi_s$ is the semiconductor's electron affinity and $\phi_m$ is the metal contact's work function.

Contacts between 2D semiconductors and conventional metals are affected by Fermi level pinning, whereby metal-induced and/or defect-induced gap states cause the Fermi energy to become pinned at the charge neutrality level.[2] As a result, even though the work functions of metals and the electron affinities of two-dimensional semiconductors are well known, the real barrier height is not simply $\chi_s - \phi_m$. Processing conditions, the number of defects in the 2D semiconductor, strain profiles, and other non-idealities make accurately predicting this value difficult. Our previous experimental estimates for nickel + monolayer $WS_2$ systems put the barrier height between 170 and 400 meV;[9] here, we expand this range to 10 to 510 meV.

We note that because we do not explicitly model Fermi level pinning or other non-idealities, e.g., spatial variations of the Schottky barrier, the barrier height we extract represents an *effective* Schottky barrier height. To gain additional physical insights, one could develop a mapping of these extracted effective barrier heights to additional physical parameters (e.g., the Fermi pinning factor). Alternatively, one could directly incorporate additional non-idealities (such as Fermi level pinning) when building the TCAD training data set, possibly using parameterized functions involving new physical parameters, as a means of studying these non-idealities in more detail. Including these factors in simulations may increase the computational cost of building these training sets, so it could also be worthwhile to pretrain networks using data with effective barrier models and then fine-tune using more comprehensive simulations.

**Density of states**: a semiconductor's effective density of states $N_C$ refers to the number of energy levels in which electrons can reside, with a larger density of states meaning that more electrons can become available to conduct current (i.e., a higher density of states means higher conductance, assuming the same mobility). The density of states can be estimated from simulation techniques, such as density functional theory; however, local strain profiles, spacing of higher-energy valleys, and other extrinsic factors can affect a material's effective density of states. Typically, $N_C \approx 5\times10^{11}$ cm$^{-2}$ for 2D semiconductors;[10] here, because of these uncertainties, we use a large expanded range of $2\times10^{11}$ to $9\times10^{12}$ cm$^{-2}$ to comfortably encapsulate the expected $N_C$.

**Donor defect profiles**: experimental studies suggest that sulfur vacancies in two-dimensional semiconductors may act as electron donors.[11] Photoluminescence studies suggest that these vacancies follow Gaussian distributions with respect to energy, with shallow defects located a few tenths of an eV below the conduction band edge and deep states located nearer to the mid gap. Since we do not probe the deep subthreshold region in this work, we consider only the shallow defect states. A previous work[11] assigns these states a median energy $E_{D,mid} \approx 0.15$ eV below the conduction band edge and an energy width (i.e., standard deviation) $\sigma_D \approx 72$ meV (from Figure 3C of ref. [11]). See **Figure 2B** in the main text for a visual depiction of this profile and these parameters. Here, we use these as rough center points for our estimated ranges, sampling $E_{D,mid}$, and $\sigma_D$ both between 20 – 200 meV. The peak concentration $N_{D0}$ for these donor states can vary tremendously depending on film quality, making it difficult to estimate based on previous works. Here, we intentionally use a large range of $3\times10^{12}$ – $3\times10^{13}$ eV$^{-1}$ cm$^{-2}$ to encapsulate reasonable values for $N_{D0}$.

**Band tail acceptor-like defect profiles**: experimentally, 2D semiconductors do not have sharp conduction band edges; instead, the density of states decreases exponentially as $E$ decreases below $E_C$. Although some works have investigated the influence of these band tail states on transport, the precise implications of band tails on electron transport in 2D semiconductors are unclear. In particular, the mobility of electrons contributed by band tail states is unclear, although it appears to be much less than that of conduction band electrons.

In this work, we adapt a convention previously applied for amorphous oxide semiconductors and model band tails as acceptor states and assign them 0 mobility.[12] This assumption is also made in the multiple-trap-and-release model, which is frequently used to describe transport in organic semiconductors.[13] Even though these states do not participate directly in conduction, their electrostatics are still considered explicitly: a higher density of tail states shifts the threshold voltage to the right and worsens subthreshold swing.

Previous work has estimated that the peak band tail density $N_{A0}$ and characteristic energy decay length $\sigma_A$ that characterize the exponential distribution [**Eq. (5)** in the main text] vary from $3\times10^{12}$ – $1.8\times10^{13}$ eV$^{-1}$ cm$^{-2}$ and from ~56 meV to 170 meV for similar 2D semiconductor systems.[14] Here, we expand these ranges to $6\times10^{11}$ – $3.7\times10^{13}$ eV$^{-1}$ cm$^{-2}$ and 50 – 300 meV, respectively, to comfortably encapsulate these initial estimates.

## Supplementary Section 3: Feature engineering and ablation study

The inverse neural network shown in **Figure 3A** of the main text aims to reverse-engineer physics-based simulation model parameters from electrical measurements. For each device, we input drain current *vs*. gate-to-source voltage ($I_d$-$V_{gs}$) curves at 32 fixed $V_{gs}$ values, with applied drain-to-source biases $V_{ds}$ = 0.1 and 1 V. For each $V_{ds}$, we also perform simple feature engineering by inputting each current in logarithmic space, along with $\partial I_d/\partial V_{gs}$ and $\partial \log_{10} I_d/\partial V_{gs}$, for a total of eight unique features per device.

In principle, model parameters can be fit from experimental or simulated data using $I_d$-$V_{gs}$ data at a single $V_{ds}$. Further, because derivatives and logarithms are implicitly contained within the data itself, seven of our eight features may seem unnecessary. Here, we justify from domain knowledge why one should expect that including these features should improve the performance of the neural network. After, we demonstrate empirically with an ablation study that failing to include these terms results in worse performance.

**Including multiple drain biases**: transistors with Schottky contacts, such as experimentally relevant two-dimensional transistors, can be conceptually thought of as an ideal ohmic transistor in series with two separate diodes that represent the source and drain.[15] These diodes are nonlinear circuit elements, meaning here that their resistances are functions of the voltage drops across them. Including multiple drain biases should help to ensure that the resistance of both the channel and the source/drain take on additional unique profiles, and thus aid in estimating the Schottky barrier height, one of our key fitting parameters.

**Including logarithmic $I_d$-$V_{gs}$ data**: The off-state of the transistor provides essential information about key physical model parameters, including the acceptor and donor profiles that we parameterize using the five separate fitting parameters in **Table 1** in the main text. However, $I_d$ in the off-state (low $V_{gs}$) can be orders of magnitude smaller than $I_d$ in the on-state. For example, in **Figure 5A** in the main text, the ratio of on- to off-state current for many of our experimental $WS_2$ transistors is $>10^5$. If we did not take the logarithm of $I_d$, differences in off-state currents would be washed out, causing us to lose essential information in the data.

**Including derivatives**: mobility is commonly extracted from transistor measurements by finding the transconductance $g_m = \partial I_d/\partial V_{gs}$ and then estimating the mobility as $\mu = g_m/(C_g V_{gs} - V_T)$, where $C_g$ is the gate capacitance. This mobility extraction is frequently made non-trivial by *contact gating*, i.e., the gate voltage modulates $R_C$, causing $g_m$ to change from extrinsic factors, such as the Schottky barrier height.[3,16] Similarly, the subthreshold swing (inversely proportional to $\partial \log_{10} I_d/\partial V_{gs}$ in the off-state) can often be correlated to defect profiles [acceptor and donor, e.g., **Eqs. (4) and (5)** in the main text]. Thus, we explicitly input the derivatives to make these parameters easier for the neural network to extract.

**Ablation study**: we empirically verify the importance of including each feature by ablating them from the input matrix, (i.e., by setting all entries in the relevant feature columns to 0), and then re-training the neural networks as before. This process allows us to examine how withholding the feature impacts the performance of the final trained network. For each ablation, we use 1,000 devices in our training and development sets combined and perform 5 different runs, using the same 5 sets of training and development data for each ablation. Afterwards, we test the performance of the neural network by reverse-engineering the fitting parameters, estimating the corresponding $I_d$-$V_{gs}$ curves for these extracted parameters, and then calculating the $R^2$ fit between the original and estimated $I_d$-$V_{gs}$ data. Here, all tests consider $I_d$-$V_{gs}$ curves at $V_{ds}$ = 0.1 and 1 V, where the $R^2$ fit is averaged in both linear and logarithmic space at both $V_{ds}$ values (same as in the main text).

As shown in **Supplementary Figure S1A**, we find that the resultant $I_d$-$V_{gs}$ fits become quite poor when we fail to include the logarithmic data or when we remove either drain bias. We find that the quality of the $I_d$-$V_{gs}$ fits remain acceptable when we do not include the derivatives of the $I_d$-$V_{gs}$ curves and when we include only the logarithmic data; however, as shown in **Supplementary Figure S1B**, the error in the fitting parameter increases if we exclude these features.

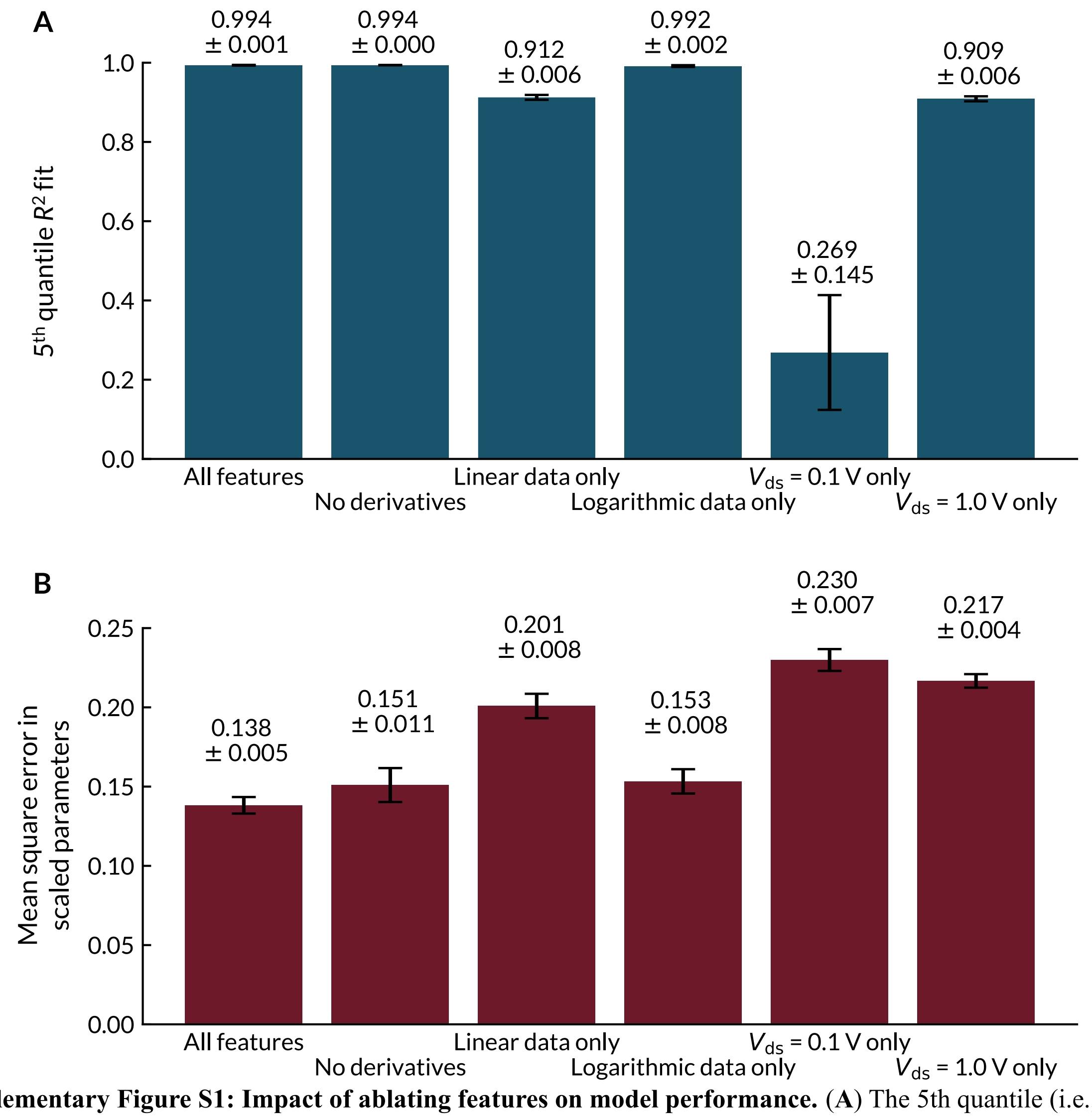

**Supplementary Figure S1: Impact of ablating features on model performance.** (**A**) The 5th quantile (i.e., worst 5%) $R^2$ fit (higher is better) and (**B**) mean square error in scaled parameters (lower is better) for different configurations of input features. We find that when removing our engineered features, the resultant fits become worse and the error in the fitting parameters increases. "Linear data only" and "logarithmic data only" denote that only the linear-scale $I_d$-$V_{gs}$ curves and their derivatives, or logarithmic-scale $I_d$-$V_{gs}$ curves and their derivatives, respectively, were included in the input matrix.

## Supplementary Section 4: Details for pre-processing, bootstrapping, and testing

### Interpolation

Our forward neural network (**Supplementary Figure S2A**) outputs $I_d$-$V_{gs}$ curves along 32 fixed $V_{gs}$ values, evenly spaced from -6 to 50 V. (Our Sentaurus Device simulations stopped just outside of the $V_{gs}$ range of -6 to 50 V; hence, we actually use a range of -5.9 to 49.9 V throughout this work.) Our inverse neural network (**Figure 3A** in the main text) accepts $I_d$-$V_{gs}$ curves and their derivatives along the same $V_{gs}$ points. In our Sentaurus Simulations, we compute drain current *vs.* gate-to-source voltage ($I_d$-$V_{gs}$) profiles across a similar range of $V_{gs}$ values; however, the exact $V_{gs}$ grid changes between simulations to maximize computational efficiency and hence does not align perfectly with the 32 fixed $V_{gs}$ values in the network's output and input layers. Therefore, we pre-process our simulated $I_d$-$V_{gs}$ data by interpolating $I_d$ along the fixed $V_{gs}$ points required by our neural networks, using linear and logarithmic interpolation to interpolate $I_d$ and $\log_{10} I_d$, respectively. This interpolation scheme and all pre-processing steps are implemented in process_devices.py in our GitHub Repository.[17]

### Noise floor

Experimentally, the lowest drain current that can be reliably measured is limited by the *noise floor* of the measurement instrument. For our experimental monolayer $WS_2$ transistor measurements, our noise floor is ~$10^{-5}$ μA/μm (**Figure 5A** in the main text). Our Sentaurus Device simulations allow us to simulate $I_d$ much less than $10^{-5}$ μA/μm; however, we do not wish to train our neural networks to rely upon features that we do not measure experimentally. Thus, we manually impose a noise floor of $5\times10^{-5}$ μA/μm on all Sentaurus-simulated data; that is, any $I_d < 5\times10^{-5}$ μA/μm is manually mapped to $5\times10^{-5}$ μA/μm. To ensure that the noise floor in our simulated training/test sets and experimental test sets are identical, we perform the same mapping on our experimental $I_d$-$V_{gs}$ curves prior to feeding them into our inverse neural network for inference.

### Data formatting and min-max scaling

The inverse neural network accepts an input matrix for each device consisting of $I_d$-$V_{gs}$ and $\log_{10}(I_d)$-$V_{gs}$ curves and their derivatives (see input in **Figure 3A**), where columns correspond to individual features [e.g., $I_d$, $\log_{10}(I_d)$, $\partial I_d/\partial V_{gs}$] and rows correspond to the 32 fixed $V_{gs}$ points. For training, these matrices for multiple devices are combined into a third order tensor $\boldsymbol{U}$. The first dimension of $\boldsymbol{U}$ corresponds to the device index, the second dimension corresponds to the 32 fixed $V_{gs}$ points, and the third dimension corresponds to an individual feature. Here, we index starting from 0, and a colon (:) denotes selecting every entry along an index. For example, the slice $\boldsymbol{U}$[j, :, 0] is the $I_d$-$V_{gs}$ curve at $V_{ds}$ = 0.1 V for the $j^{th}$ device.

We prepare the scaled tensor $\tilde{U}$ by min-max scaling $\boldsymbol{U}$ feature-wise between -1 and 1. Each $i^{th}$ matrix slice along the last dimension of $\boldsymbol{U}$ is scaled independently using the minimum and maximum values across that individual slice. That is, for each $i^{th}$ slice along the third dimension,

$$\tilde{U}[:,:,i] = 2\left(\frac{U[:,:,i] - \min(U[:,:,i])}{\max(U[:,:,i]) - \min(U[:,:,i])}\right) - 1$$

Next, we consider the output of the forward neural network, $\boldsymbol{V}$, which is also a third-order tensor. $\boldsymbol{V}$ is a subset of $\boldsymbol{U}$ [$\boldsymbol{U}$ contains $I_d$-$V_{gs}$ and $\log_{10}(I_d)$-$V_{gs}$ curves and their derivatives at two drain biases; $\boldsymbol{V}$ contains the $I_d$-$V_{gs}$ and $\log_{10}(I_d)$-$V_{gs}$ curves, but not their derivatives]. Thus, $\boldsymbol{V}$ and its scaled equivalent $\tilde{V}$ are built simply by taking the required slices from $\boldsymbol{U}$ and $\tilde{U}$.

For the $i^{th}$ device, the input of the forward neural network and output of the inverse neural network is the vector $\boldsymbol{y^{(i)}}$. For training, these vectors are stacked into the matrix $\boldsymbol{Y}$, where the first dimension (rows) corresponds to the device index, and the second dimension (columns) corresponds to the fitting parameter. For example, $\boldsymbol{Y}[i,j]$ corresponds to the $j^{th}$ fitting parameter of the $i^{th}$ device.

We min-max scale $\boldsymbol{Y}$ between -1 and 1 for each fitting parameter:

$$\tilde{Y}[:,j] = 2\left(\frac{Y[:,j]-\min(Y[:,j])}{\max(Y[:,j])-\min(Y[:,j])}\right)-1$$

All min-max scaling is implemented in process_devices.py in our GitHub Repository.

## Calculating the current mismatch in the inverse neural network loss function

In **Eq. (6)** of the main text, we introduce the term $E_{I_d}^{(i)}$ to allow the loss function to penalize errors in the predicted $I_d$-$V_{gs}$ curves for the $i^{th}$ device. Here, $E_{I_d}^{(i)}$ is the sum of three terms that individually quantify errors in the $I_d$-$V_{gs}$ curves, their slopes, and their curvatures:

$$E_{I_d}^{(i)} = {L_{I_d}^{(i)}}^2 + {L_{\Delta I_d}^{(i)}}^2 + {L_{\Delta^2 I_d}^{(i)}}^2 \tag{S1}$$

where:

$${L_{I_d}^{(i)}}^2 = \frac{1}{2N_{V_{ds}}N_{V_{gs}}}\sum_j^{N_{V_{ds}}}\sum_k^{N_{V_{gs}}}\left(\left(I_{true}^{(i,j,k)}-I_{pred}^{(i,j,k)}\right)^2 + \left(\log_{10} I_{true}^{(i,j,k)} - \log_{10} I_{pred}^{(i,j,k)}\right)^2\right) \tag{S2}$$

$${L_{\Delta I_d}^{(i)}}^2 = \frac{1}{2N_{V_{ds}}\left(N_{V_{gs}}-1\right)}\sum_j^{N_{V_{ds}}}\sum_k^{N_{V_{gs}}-1}\left(\left(\Delta I_{true}^{(i,j,k)}-\Delta I_{pred}^{(i,j,k)}\right)^2 + \left(\Delta \log_{10} I_{true}^{(i,j,k)} - \Delta \log_{10} I_{pred}^{(i,j,k)}\right)^2\right) \tag{S3}$$

$${L_{\Delta^2 I_d}^{(i)}}^2 = \frac{1}{2N_{V_{ds}}\left(N_{V_{gs}}-2\right)}\sum_j^{N_{V_{ds}}}\sum_k^{N_{V_{gs}}-2}\left(\left(\Delta^2 I_{true}^{(i,j,k)}-\Delta^2 I_{pred}^{(i,j,k)}\right)^2 + \left(\Delta^2 \log_{10} I_{true}^{(i,j,k)} - \Delta^2 \log_{10} I_{pred}^{(i,j,k)}\right)^2\right) \tag{S4}$$

where $N_{V_{ds}}$ and $N_{V_{gs}}$ are the number of $V_{ds}$ and $V_{gs}$ values, respectively, and $j$ and $k$ denote the $j^{th}$ $V_{ds}$ and $k^{th}$ $V_{gs}$ value, respectively. $\Delta$ and $\Delta^2$ denote first and second differences with respect to $V_{gs}$, e.g.,

$$\Delta I_{true}^{(i,j,k)} = I_{true}^{(i,j,k)} - I_{true}^{(i,j,k-1)} \tag{S5}$$

$$\Delta^2 I_{true}^{(i,j,k)} = \Delta I_{true}^{(i,j,k)} - \Delta I_{true}^{(i,j,k-1)} \tag{S6}$$

## Bootstrapping studies

We use a bootstrapping approach in **Figures 4A,B** in the main text to understand how varying the training set size influences the neural network's performance when reverse-engineering two-dimensional transistors. For each bootstrapped simulation, we build smaller training subsets by randomly selecting the specified number of devices from the original $I_d$-$V_{gs}$ simulations. Afterwards, we initialize the forward and inverse models with random weights using the Glorot uniform initializer as implemented in TensorFlow, and carry out the training scheme using these smaller training sets.

## $R^2$ calculations

Throughout the main text and supplementary information, we report individual $R^2$ values to measure the goodness of fit between data sets containing multiple baseline truths and predictions. A conventional $R^2$ that summarizes multiple data sets is not typically well defined. Here, for a set of $N$ curves, we calculate the individual $R^2$ for every curve and then take the average of these values. For our $I_d$-$V_{gs}$ curves, we include both the $R^2$ in linear space (i.e., as normally calculated) and $R^2$ in logarithmic space (i.e., transforming both the baseline truth and predicted $I_d$ by taking their base 10 logarithm and then calculating the $R^2$ with these transformed values) in the above sum. These $R^2$ values are calculated on baseline and predicted $I_d$-$V_{gs}$ curves after interpolating them onto 32 evenly spaced $V_{gs}$ points and after imposing noise floor mapping, as described in the "interpolation" and "noise floor" procedures earlier in this section.

## Supplementary Section 5: Details for the forward neural network

Here, we discuss the *forward neural network* (architecture in **Supplementary Figure S2A**) used to predict $I_d$-$V_{gs}$ curves from the model parameters in **Table 1**. In essence, this neural network approximates the original physics-based TCAD solver or, equivalently, it approximates $\boldsymbol{M}_{\mathbf{forward}}$ in **Eq. (1)**. We emphasize that the forward neural network should not be confused with the *inverse neural network* that we use to reverse-engineer model parameters; the forward neural network is a surrogate model that (i) provides an augmented data set for pretraining our inverse neural network and (ii) is called when evaluating the loss function of our inverse neural network using a tandem approach (as in **Figure 3B** in the main text).

After scaling all data, we train the forward neural network to output $I_d$-$V_{gs}$ curves with $V_{gs}$ values evenly spaced from -6 to +50 V (see **Supplementary Section 4** for scaling and preprocessing procedures). The network outputs $I_d$-$V_{gs}$ curves at $V_{ds}$ = 0.1 and 1.0 V, in both linear and logarithmic space, yielding four separate curves. Although predicting both $I_d$ and $\log_{10}(I_d)$ explicitly may seem redundant, we do so because the hyperbolic tangent activation functions in the network's output layer have range (-1, 1) and hence cannot easily predict outputs that span multiple orders of magnitude. Thus, we explicitly output $\log_{10}(I_d)$ to ensure we can accurately predict the off-state current, which tends to be many orders of magnitude smaller than the on-state current.

### Forward neural network loss function

We use the loss function shown in **Eq. (S7)** when training the forward neural network. Here, we aim to ensure that the forward neural network predicts both the linear and $\log_{10} I_d$ values, along with the slopes and curvatures of the linear and $\log_{10} I_d$-$V_{gs}$ curves, across all $V_{ds}$ values.

$$\mathrm{L_{forward}} = \frac{1}{\mathrm{N_{devices}}} \sum_{\mathrm{i}}^{\mathrm{N_{devices}}} \left( \mathrm{L}_{\mathrm{I_d}}^{(\mathrm{i})\,2} + \mathrm{L}_{\Delta \mathrm{I_d}}^{(\mathrm{i})\,2} + \mathrm{L}_{\Delta^2 \mathrm{I_d}}^{(\mathrm{i})\,2} \right) \tag{S7}$$

Here, $N_{devices}$ is the number of devices in the training set and $i$ denotes the $i^{th}$ device. Other terms in **Eq. (S7)** quantify the error in the $I_d$-$V_{gs}$ measurements, along with errors in their slopes and curvatures, and are defined in **Eqs. (S2) to (S4)**. Note that, unlike in **Eq. (6)** in the main text, we do not take the square root of the bracketed term in the forward neural network's loss function.

### Forward neural network performance

Because physics-based (TCAD) $I_d$-$V_{gs}$ training data are computationally expensive, we wish to obtain high quality fits using as little training data as possible. Thus, we assess the forward network's performance *vs*. the number of devices included in the training set using the bootstrap procedure described in **Supplementary Section 4.** Here, we find that for all training set sizes considered, our forward neural network can achieve a median $R^2 \geq 0.98$ on the test set (**Supplementary Figure S2B**), indicating a good match between the original simulated current and the approximation from the forward neural network. We are also interested in the network's performance across the vast majority of the test set; thus, we plot the 5$^{th}$ quantile $R^2$ (i.e., the $R^2$ corresponding to the worst 5% of fits) on the same axes, and we plot sample cumulative histograms of $R^2$ for different training set sizes in **Supplementary Figure S2C**. We find that the worst 5% of the fits still match well when the training set contains at least ~500 devices, for which we achieve a 5$^{th}$ quantile $R^2 \approx 0.94$. We show sample fits corresponding to the median $R^2$ after training on 250, 1,000, and 4,000 devices in **Supplementary Figures S2D-F** to confirm agreement between the original physics-based simulation and the network's predicted $I_d$.

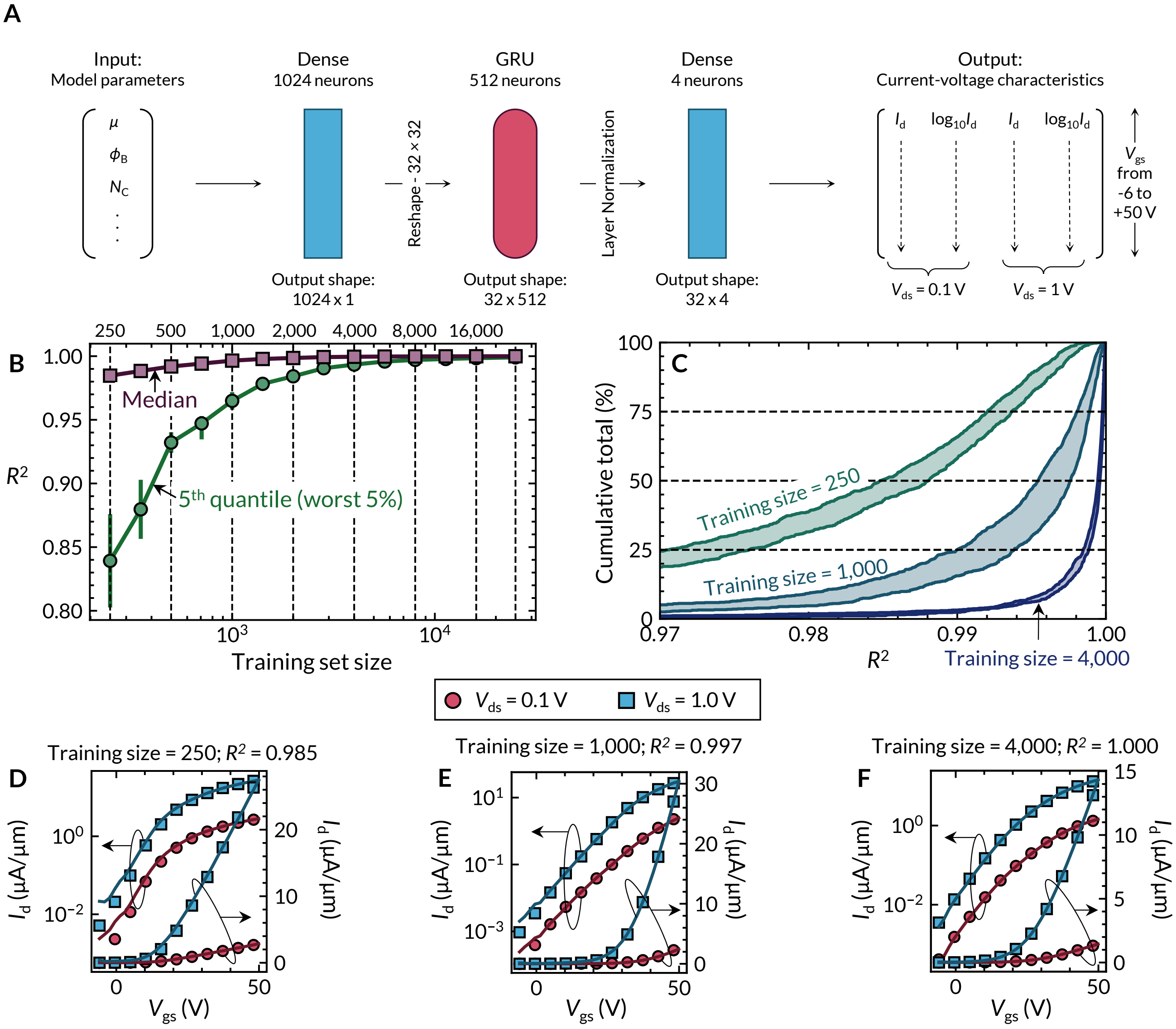

**Supplementary Figure S2: Performance of the forward neural network used as a surrogate model**. (**A**) The architecture of the neural network used to approximate the physics-based current model ("the forward neural network"). Dense layers use the hyperbolic tangent (tanh) activation function. The gated recurrent unit (GRU) layer uses tanh for its activation function and sigmoid for its recurrent activation function. (**B**) Median and 5th quantile (i.e., worst 5%) $R^2$ fits between the $I_d$ calculated from the physics-based simulation and estimated by the forward neural network *vs.* the number of devices in the training set. Model fits were evaluated using a test set of 1,000 devices that were unseen by the model during training and network optimization. Error bars show standard deviation across five bootstrap tests (see **Supplementary Section 4** for details); error bars are often extremely small and hence are not always visible. (**C**) Cumulative $R^2$ fits for various training set sizes, where shaded regions show the range for the cumulative $R^2$ across the three bootstrapped tests. Current *vs*. gate-to-source voltage plots corresponding to the median $R^2$ fits after training on (**D**) 250 devices, (**E**) 1,000 devices, and (**F**) 4,000 devices (each plotted for the worst-performing bootstrapped network). Note that each device shown is different, because they are generated from a distribution of input parameters as listed in **Table 1**. Symbols: original physics-based simulations; lines: neural network predictions. The data are shown on both logarithmic (left) and linear axes (right), at $V_{ds}$ = 0.1 V (red symbols) and 1.0 V (blue symbols).

To assess the performance of the inverse neural network on the Sentaurus-simulated test set in **Figure 4** of the main text and during our ablation study in **Supplementary Section 3**, we compare the original $I_d$-$V_{gs}$ curves to those obtained using the reverse-engineered parameters. To estimate the reverse-engineered $I_d$-$V_{gs}$ curves, we take these extracted parameters and input them into a well-trained forward neural network. This forward network was trained on all 25,000 devices in our original physics-based training set and followed the same training procedure as described in **Supplementary Section 4**, with additional epochs and annealing steps to further minimize errors.

We test this well-trained forward network on the same test set of 1,000 devices and find that it reproduces the original Sentaurus-simulated $I_d$-$V_{gs}$ characteristics exceptionally well: as shown in **Supplementary Figure S3**, this forward neural network achieves a $0.5^{th}$ quantile, $1^{st}$ quantile, and $2.5^{th}$ quantile (i.e., worst 0.5%, worst 1%, and worst 2.5%) $R^2$ fit of 0.9915, 0.9970, and 0.9973, respectively, validating the use of this method for estimating $R^2$.

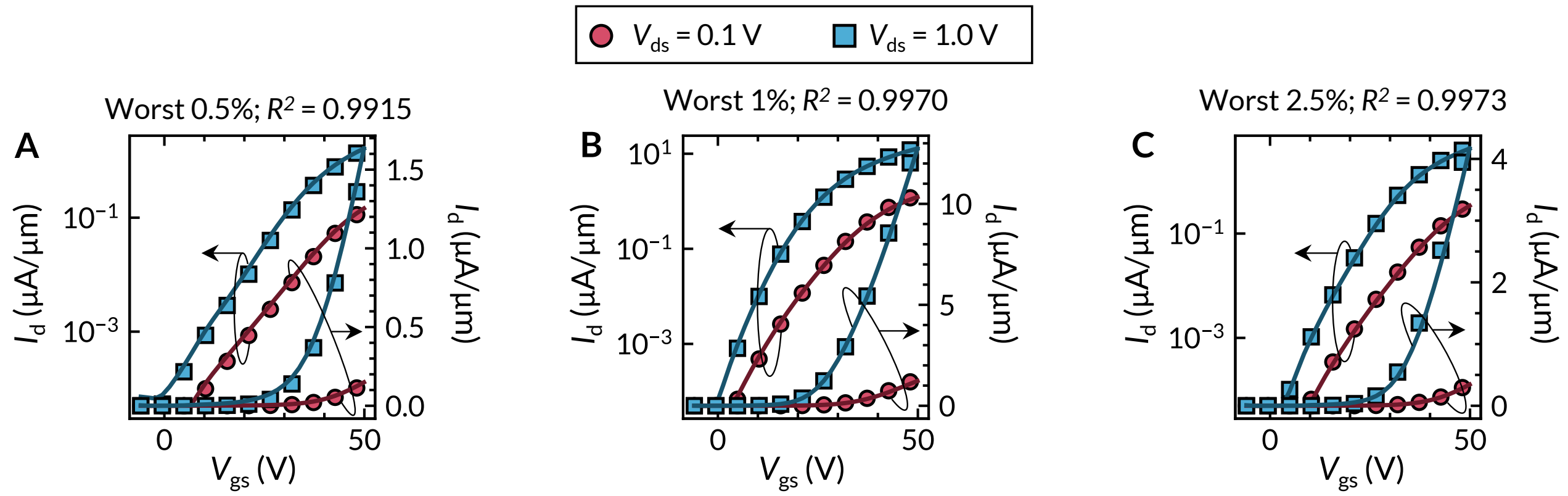

**Supplementary Figure S3: Performance of the well-trained forward neural network used to estimate $R^2$ in Figure 4 of the main text**. Current *vs*. gate-to-source voltage plots of the worst (**A**) 0.5%, (**B**) 1%, and (**C**) 2.5% $R^2$ sets obtained when testing the well-trained forward neural network on the Sentaurus-simulated test set. Even for these pessimistic scenarios, we find high $R^2 > 0.99$ across nearly all of the test set. Symbols: original physics-based simulations; lines: neural network predictions. The data are shown on both logarithmic (left) and linear axes (right), at $V_{ds} = 0.1$ V (red symbols) and 1.0 V (blue symbols).

## Supplementary Section 6: Complete parameter extractions

In **Supplementary Figure S4**, we show the actual *vs*. predicted values and corresponding error histograms for the eight model parameters from **Table 1** in the main text. We extract these parameters for the test set of 1,000 Sentaurus-simulated devices using the same inverse neural network as used in **Figures 4C-H** in the main text (i.e., this neural network was trained using 500 devices included in the main physics-based training set). To avoid cluttering plots of actual *vs*. predicted data, we plot only 250 out of 1,000 points from the test set when plotting actual *vs*. predicted values.

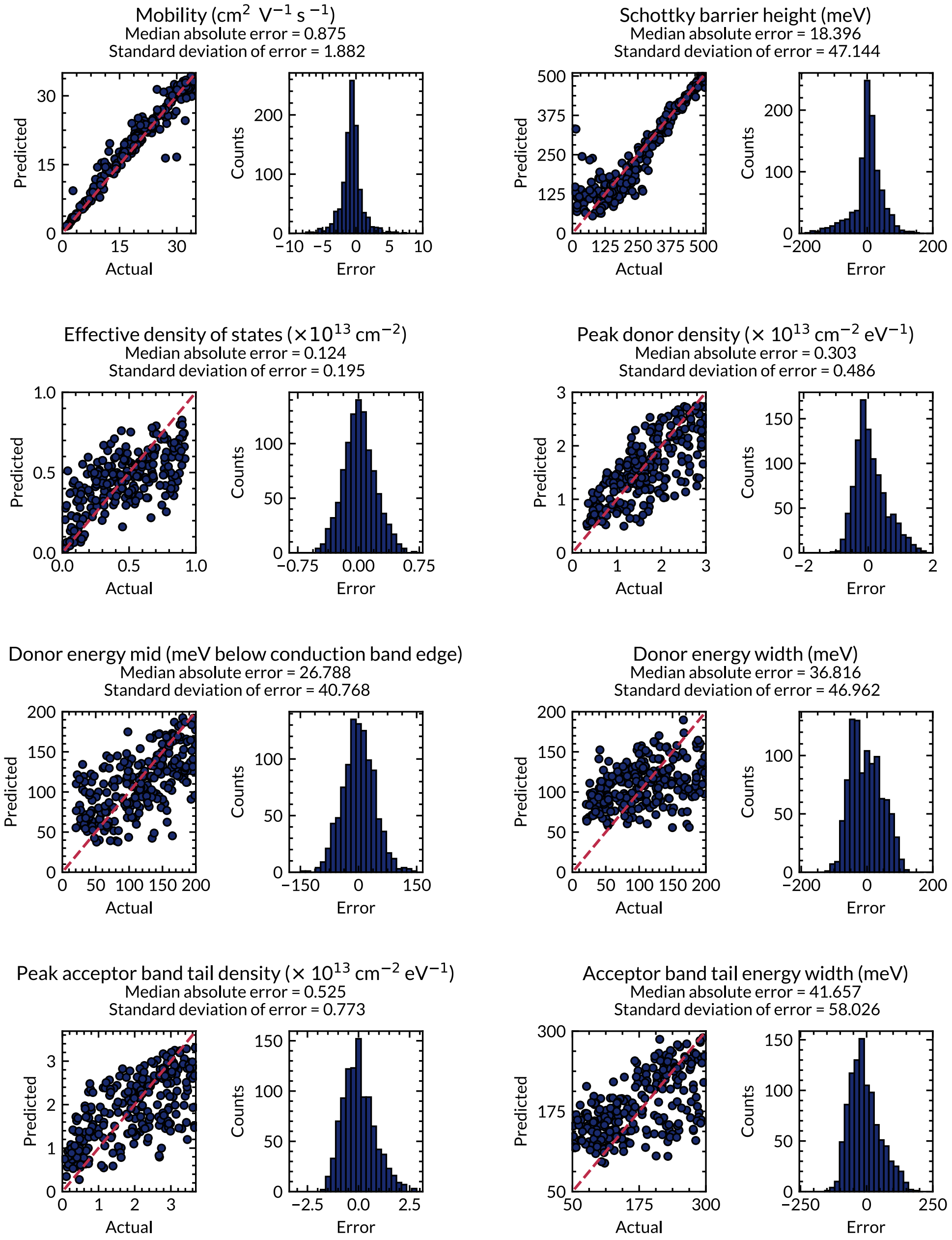

**Supplementary Figure S4: Errors for individual parameters extracted from the simulated test set discussed in the main text.** Each subpanel shows actual *vs*. predicted values and corresponding error histograms for the eight model parameters from **Table 1**. The median absolute errors and standard deviation of errors were calculated from the full 1,000 devices, and the histograms show data from all devices as well. All units for each plot, except for histogram counts, are as indicated in the corresponding subpanel titles. We truncate the *x*-axes of the histograms at ± ~4×the standard deviation of error for clarity.

When training new deep learning models for parameter extraction, we strongly recommend preparing similar statistical plots as in **Supplementary Figure S4** using synthetic test data to assess which parameters are especially error-prone. Once these plots are prepared, the accuracy of extracted parameters for an individual device can be estimated by examining the range of errors given the extracted values. For instance, if the deep learning approach estimates a Schottky barrier height $\phi_B \approx 500$ meV, the $\phi_B$ statistics from **Supplementary Figure S4** suggest that this estimate is likely quite accurate. Alternatively, if $\phi_B$ is estimated to be ~125 meV, then this figure suggests that the extracted value is more error-prone, likely being somewhere between ~0 and ~200 meV. This procedure could be repeated for every device parameter to achieve detailed distributions (e.g., histograms) estimating their uncertainty.

Another way to estimate the errors in extracted parameters for a particular device is to train multiple inverse neural networks from randomly initialized starting points. Then, one could use each of the bootstrapped networks to extract the parameter set for the same device multiple times. The value of each extracted parameter will vary across some range, with a narrow range indicating higher reliability. We note that the entire training procedure we implement here, including training the forward neural network, pretraining the inverse neural network, and fine-tuning the inverse neural network, takes less than 15 minutes on a single NVIDIA RTX 4080 graphics processing unit, making this analysis computationally tractable.

To examine how interplay between parameters could affect extraction errors, we examine the distribution of errors in the peak band tail density $N_{A0}$ and acceptor band tail energy width $\sigma_A$ as a test case. Together, these parameters define the acceptor density $N_A$ *vs.* energy $E$ profile, $N_A(E)$. From **Equation 5** in the main text, $N_A(E)$ increases with $N_{A0}$ and decreases with $\sigma_A$; therefore, if one were to overestimate $N_{A0}$, then the overall $N_A(E)$ profile (the actual physical quantity of interest) could be partially recovered by decreasing $\sigma_A$ to compensate, or vice-versa.

To confirm that our inverse neural networks do compensate for parameter errors in this way, we plot in **Supplementary Figure S5** the difference between the actual and predicted $\sigma_A$ as a function of the error in $N_{A0}$, i.e., we plot $\sigma_A^{(actual)}$-$\sigma_A^{(pred)}$ *vs.* $N_{A0}^{(actual)}$-$N_{A0}^{(pred)}$ where the superscripts "actual" and "pred" denote actual and predicted quantities, respectively. Here, we find that there is indeed a clear trend between these errors; furthermore $N_{A0}$ overestimation is accompanied by $\sigma_A$ underestimation, or vice-versa, 77% of the time.

We note that there is likely similar interplay between all parameters; studying this interplay in detail would necessitate a thorough multivariate analysis that is beyond the scope of this work. Nevertheless, the correlation between errors shown in **Supplementary Figure S5** supports our hypothesis that some of the fitting uncertainty can be explained by interplay between parameters in our electrical simulations.

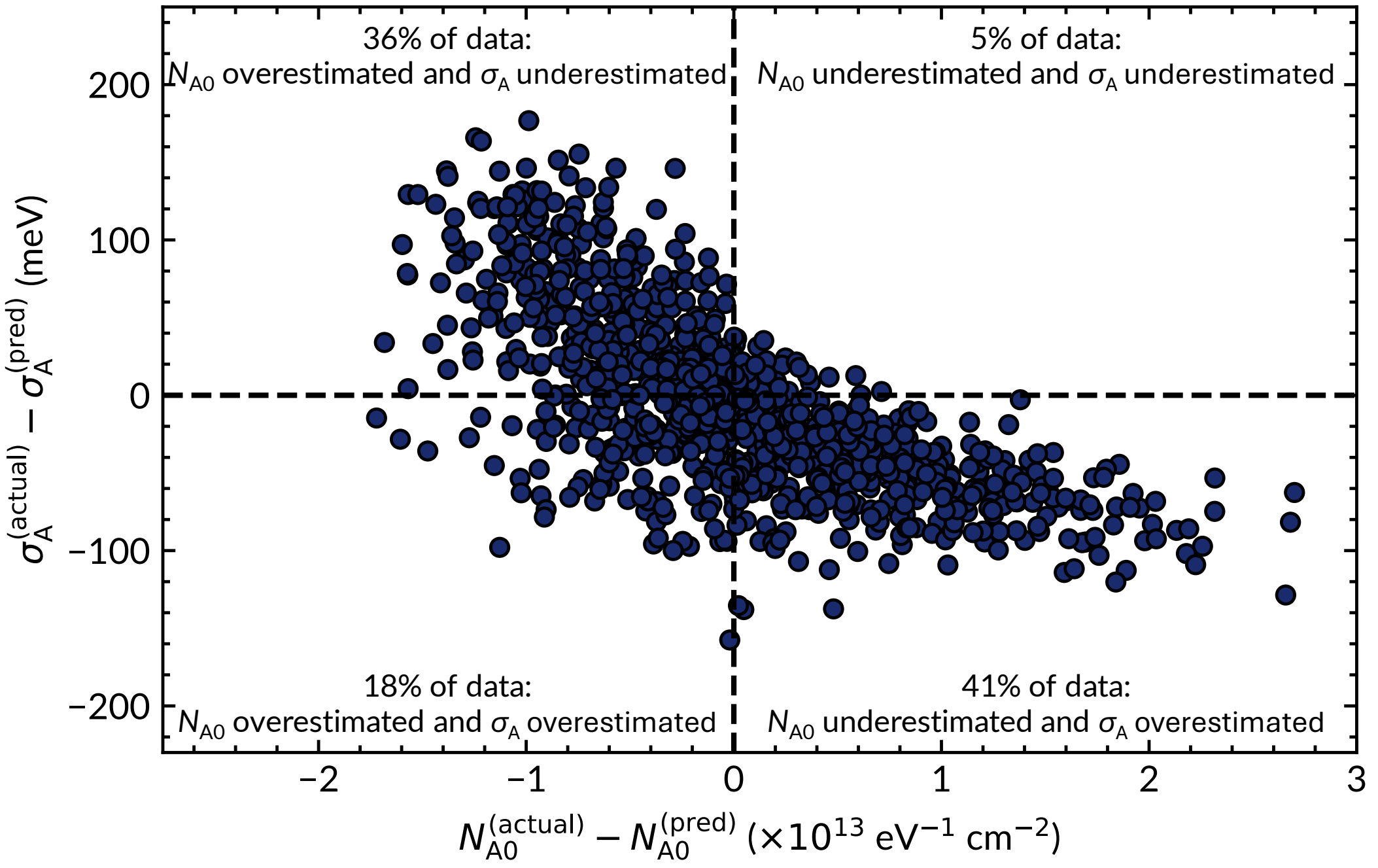

**Supplementary Figure S5: Correlation between errors in parameters that define the total band-tail-like acceptor density**. Here, we plot the error in the acceptor band tail energy width $\sigma_A$ as a function of the error in the peak acceptor band tail density $N_{A0}$, i.e., $\sigma_A^{(actual)}$-$\sigma_A^{(pred)}$ vs. $N_{A0}^{(actual)}$-$N_{A0}^{(pred)}$, where the superscripts "actual" and "pred" denote actual and predicted quantities, respectively.

## Supplementary Section 7: Fits for all experimental devices

Below, we show fits across the experimental test set of 51 monolayer $WS_2$ transistors, ordered by ascending $R^2$. We title each plot with the number of the device (ordered by the quality of the fit; 1/51 indicates the worst $R^2$ fit and 51/51 indicates the best $R^2$ fit). Filled markers and solid lines show the data on the linear scale (right) $y$-axis, and hollow markers and dashed lines show the data on the logarithmic scale (left) $y$-axis.

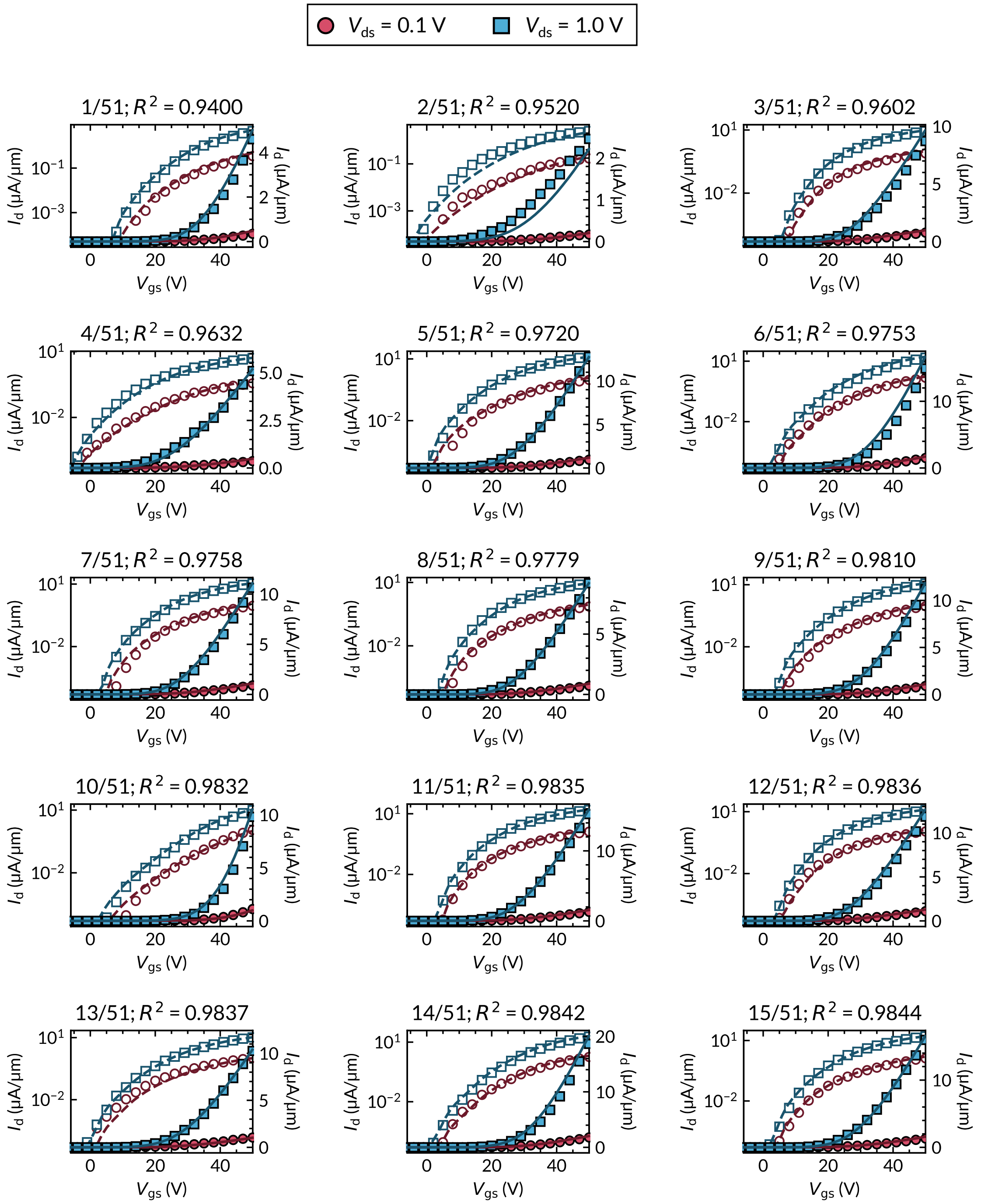

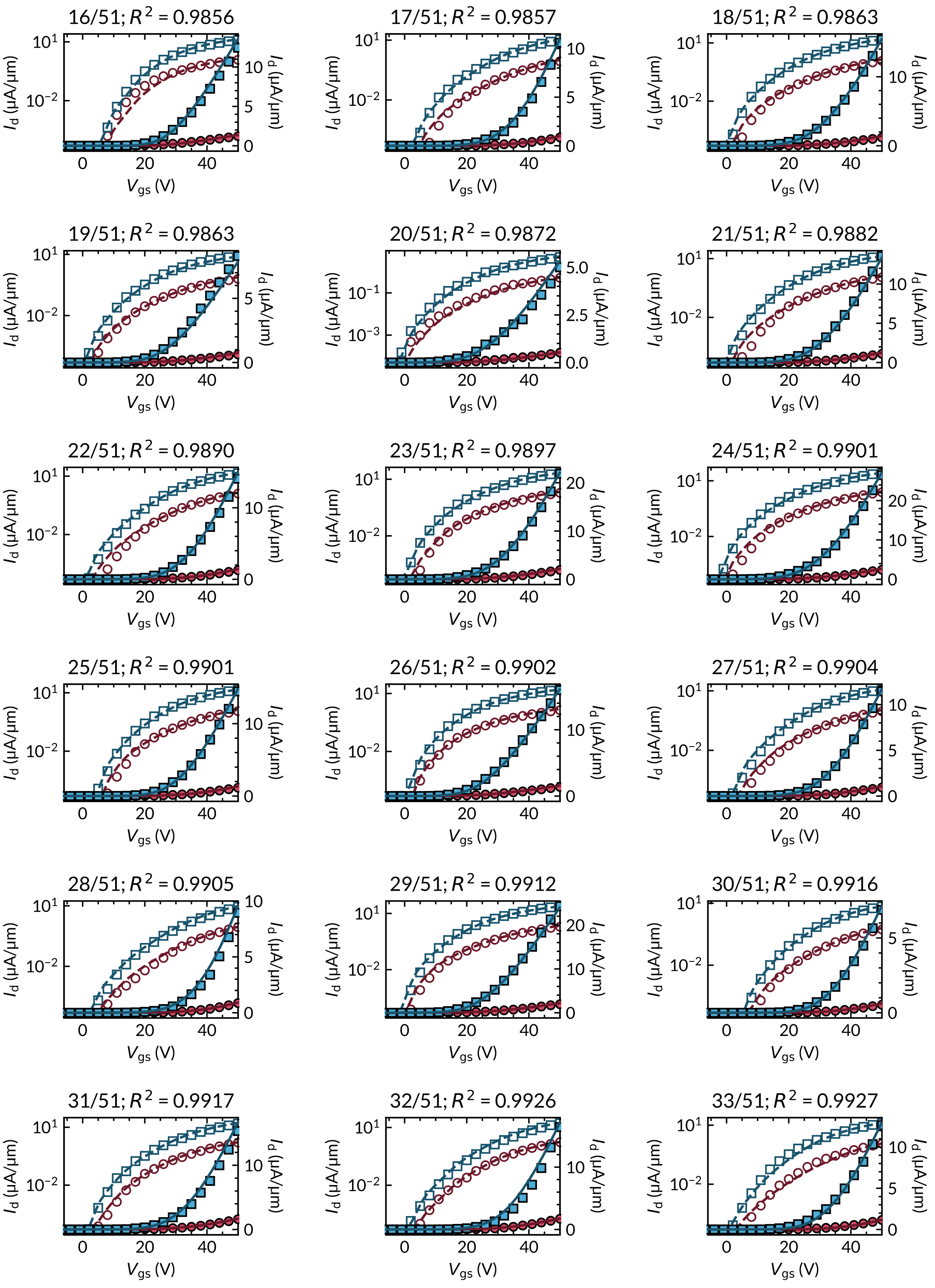
16/51; $R^2$ = 0.9856
17/51; $R^2$ = 0.9857
18/51; $R^2$ = 0.9863
19/51; $R^2$ = 0.9863
20/51; $R^2$ = 0.9872
21/51; $R^2$ = 0.9882
22/51; $R^2$ = 0.9890
23/51; $R^2$ = 0.9897
24/51; $R^2$ = 0.9901
25/51; $R^2$ = 0.9901
26/51; $R^2$ = 0.9902
27/51; $R^2$ = 0.9904
28/51; $R^2$ = 0.9905
29/51; $R^2$ = 0.9912
30/51; $R^2$ = 0.9916
31/51; $R^2$ = 0.9917
32/51; $R^2$ = 0.9926
33/51; $R^2$ = 0.9927
$I_d$ (μA/μm)
$V_{gs}$ (V)

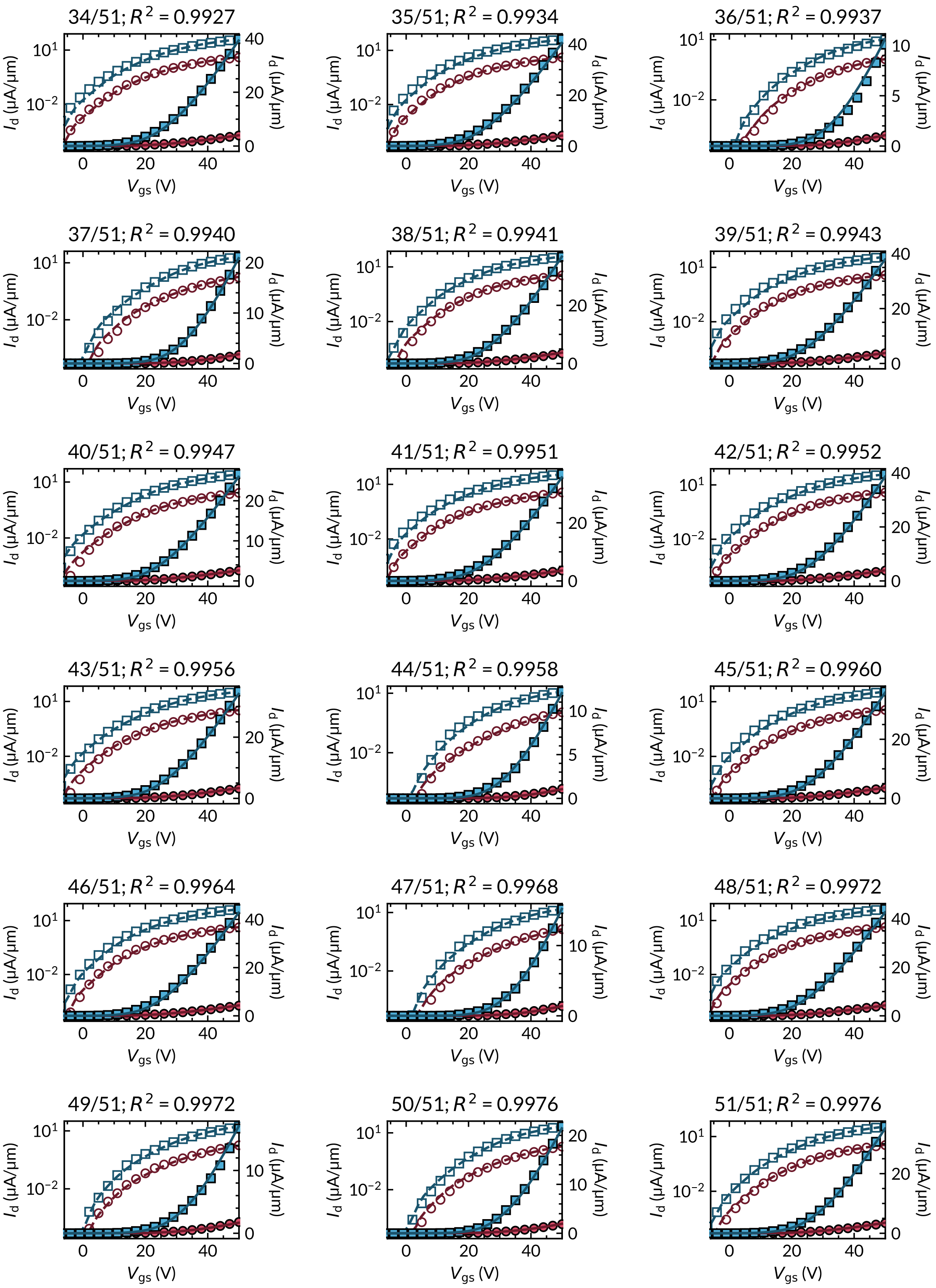
34/51; $R^2$ = 0.9927
35/51; $R^2$ = 0.9934
36/51; $R^2$ = 0.9937
37/51; $R^2$ = 0.9940
38/51; $R^2$ = 0.9941
39/51; $R^2$ = 0.9943
40/51; $R^2$ = 0.9947
41/51; $R^2$ = 0.9951
42/51; $R^2$ = 0.9952
43/51; $R^2$ = 0.9956
44/51; $R^2$ = 0.9958
45/51; $R^2$ = 0.9960
46/51; $R^2$ = 0.9964
47/51; $R^2$ = 0.9968
48/51; $R^2$ = 0.9972
49/51; $R^2$ = 0.9972
50/51; $R^2$ = 0.9976
51/51; $R^2$ = 0.9976
$I_d$ (μA/μm)
$V_{gs}$ (V)

## Supplementary Section 8: Details for high electron mobility transistor simulations

We use the fitting parameters and ranges shown in **Supplementary Table S1** for high-electron-mobility transistors (HEMTs) with 150 nm channel lengths. We build our training/test sets from the ASM-HEMT compact model,[18] and we pre-process in the same way as for two-dimensional transistors. Our training and test data sets consist of $I_d$-$V_{gs}$ curves from $V_{gs}$ = -3.5 to 0.5 V (evenly spaced across 32 values) at fixed drain biases of $V_{ds}$ = -1, 0.1, 1, 5, 10, 15, and 20 V. All neural network architectures are nearly identical to those used for two-dimensional transistors, except that the forward neural network uses 512 and 256 neurons in its first dense layer and its gated recurrent unit (GRU) layer, respectively, and the number of neurons in the input and output layers are updated to accommodate the number of fitting parameters and $V_{ds}$ values. Here, we wish to understand the impact of increasing the number of fitting parameters on the accuracy of the inverse design procedure. Thus, we bootstrap not only the training data, but also the fitting parameters themselves.

When varying the number of fitting parameters used, we select the parameters that we vary at random from **Supplementary Table S1**; all other parameters are assigned default values at random from their respective ranges. This parameter set is very similar to that used in a previous work,[19] with the ranges of some parameters having been expanded. We note that different fitting parameters could be easier or more difficult to fit compared to others. Therefore, for each of our five bootstrapped runs, we always progressively add additional fitting parameters. For example, in a bootstrap run, we begin by fitting 5 randomly selected parameters. When we fit 10 parameters, we select 5 additional parameters to vary alongside the first 5. When we fit 15 parameters, we select 5 additional parameters to vary alongside the first 10, and so on. For the next bootstrap run, we pick 5 new random fitting parameters and begin this process anew.

**Supplementary Table S1: Fitting parameters and ranges across which they vary for the ASM-HEMT compact model simulations.** The parameters and ranges across which they vary are very similar to those in a previous work[19], with some ranges expanded (parameters not listed here take on the default values from this same reference). All parameter names and units are identical to those in the ASM-HEMT model. All fitting parameters were sampled uniformly across their respective ranges. A superscripted asterisk (*) after the parameter name denotes that this sampling was performed in logarithmic (base 10) space, e.g., for a parameter with lower and upper bounds $A$ and $B$, we drew a random number $c$ sampled uniformly between $\log_{10}A$ and $\log_{10}B$ and assigned the parameter a value $y = 10^c$.

| Parameter name | Range | Parameter name | Range |
|---|---|---|---|
| cdscd | 0.01 – 0.15 | rnjgd | 15 – 30 |
| eta0 | 0.01 – 0.1 | rnjgs | 5 – 15 |
| delta | 2 – 5 | rsc | $10^{-5}$ – $1.5\times10^{-3}$ |
| gdsmin* | $10^{-12}$ – $10^{-6}$ | rth0 | 15 – 45 |
| igddio | 7.5 – 15 | thesat | 1 – 4 |
| igsdio | 2.5 – 10 | u0 | 0.15 – 0.3 |
| imin* | $10^{-15}$ – $10^{-12}$ | u0accd | 0.05 – 0.25 |
| lambda | $10^{-4}$ – $5\times10^{-3}$ | u0accs | 0.05 – 0.25 |
| mexpaccs | 1 – 5 | ua | $10^{-8}$ – $5\times10^{-7}$ |
| mexpaccd | 1 – 5 | ub* | $10^{-21}$ – $10^{-15}$ |
| nfactor | 0.2 – 0.5 | ute | -1 – -0.1 |
| njgd | 2.5 – 20 | uted | -17.5 – -5 |
| njgs | 2.5 – 20 | utes | -17.5 – -5 |
| ns0accd* | $5\times10^{15}$ – $5\times10^{20}$ | vdscale | 2 – 6 |
| ns0accs* | $5\times10^{15}$ – $5\times10^{20}$ | voff | -2.1 – -1.9 |
| rdc | $10^{-4}$ – $1.5\times10^{-3}$ | vsat | $1.5\times10^{6}$ – $2.5\times10^{6}$ |
| rigddio | $10^{-8}$ – $10^{-7}$ | vsataccs | $10^{4}$ – $1.5\times10^{5}$ |
| rigsdio | $10^{-8}$ – $10^{-7}$ | | |

We plot the 5th quantile (worst 5%), 10th quantile (worst 10%), and median $R^2$ fit as functions of the number of fitting parameters in **Supplementary Figures S6A-C** for various training size sets. Each test set contains sets of $I_d$-$V_{gs}$ curves from 1,000 HEMTs that were unseen by the neural network during training and optimization. Here, we find that increasing the number of fitting parameters makes the extraction process more error-prone if the size of the training set remains constant. This is an unsurprising consequence of the "curse of dimensionality,"[20] since the available parameter space grows exponentially as the number of fitting parameters increases, making it increasingly difficult for the neural network to fit multiple parameters simultaneously. However, as we increase the size of our training sets, we find that we can somewhat compensate for this added complexity – for example, from **Supplementary Figure S6B**, we observe that when fitting 15 parameters simultaneously, training our neural networks on 1,000 devices in our training set yields a 10th quantile $R^2 > 0.995$ on the test set. We achieve a similar $R^2$ when fitting 25 and 35 parameters after training on 4,000 and 16,000 devices, respectively (sample fits shown in **Supplementary Figures S6D-F**). In other words, within this range, we must roughly quadruple the size of the training set for every 10 additional fitting parameters to maintain the same level of accuracy.

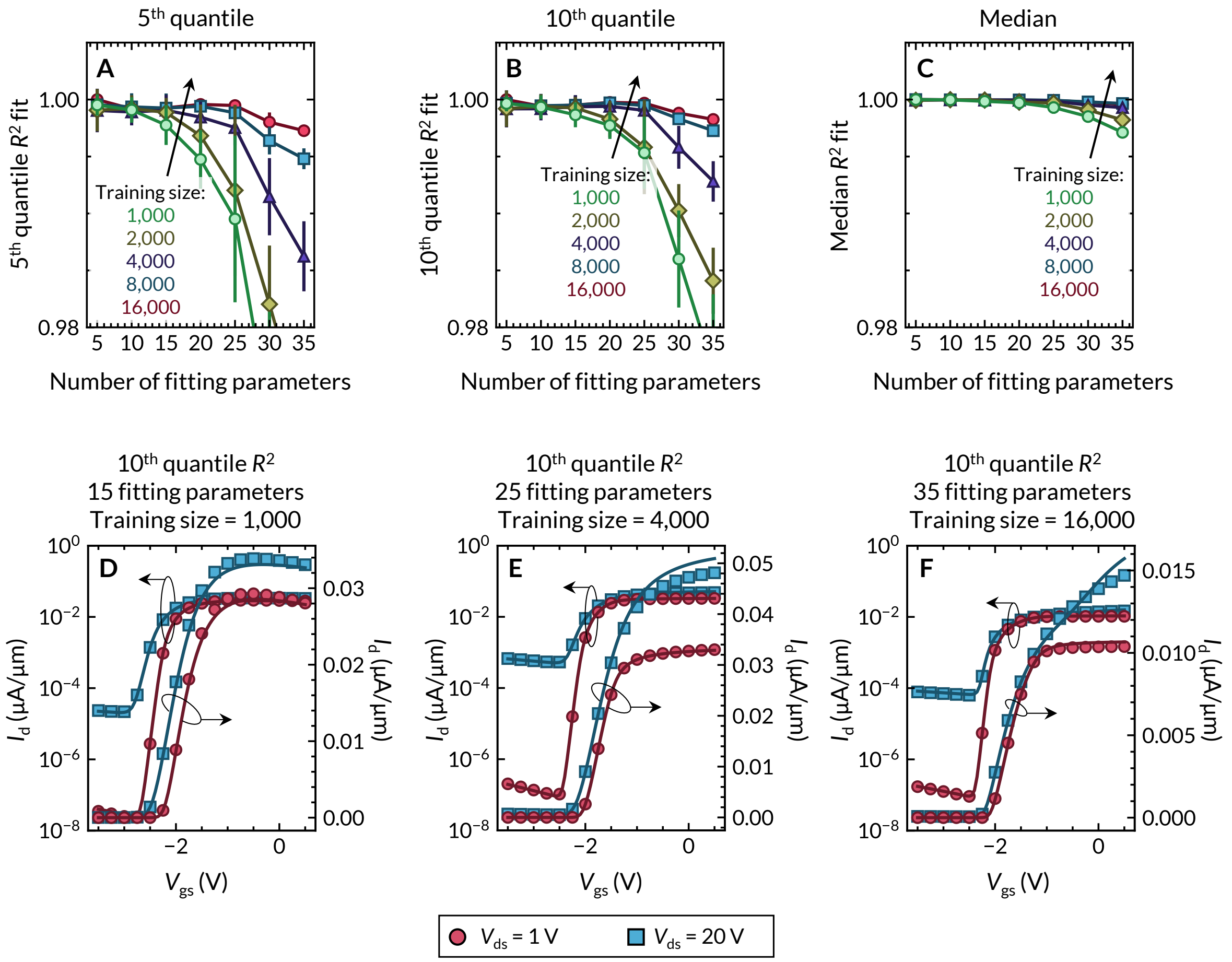

**Supplementary Figure S6: Neural network performance *vs.* complexity of fitted models.** Plots of (**A**) the 5th quantile $R^2$ fit, (**B**) the 10th quantile $R^2$ fit, and (**C**) the median $R^2$ fit plotted *vs.* the number of fitting parameters in the ASM-HEMT compact model. We show curves of the $R^2$ fit *vs.* number of fitting parameters as level curves for each training size tested. Markers and error bars show the means and standard deviations of the $R^2$ fit for five bootstrapped runs. We truncate the y-axis at $R^2 = 0.98$ to avoid overcrowding the plot. Panels (**D-F**) show the fits corresponding to the 10th quantile (worst 10%) $R^2$ from a bootstrapped model when fitting 15, 25, and 35 parameters simultaneously after training on data from 1,000, 4,000, and 16,000 devices, respectively. In all panels, we calculate $R^2$ using the procedure described in **Supplementary Section 4** across all $V_{ds}$ values used; in panels d-f, we plot $I_d$-$V_{gs}$ curves at only $V_{ds} = 1$ and 20 V to avoid cluttering the plots.

To understand how increasing the number of fitting parameters affects the extraction accuracy, we tabulate the median absolute error for each parameter across a sample bootstrap trial in **Supplementary Table S2**. For this comparison, we consider a fixed training set size of 16,000.

As expected, the error for each individual fitting parameters tends to increase with the number of parameters (i.e., moving left-to-right across a row in **Supplementary Table S2**). This occurs because the overall size of the parameter space increases, making fitting more difficult. Although the extraction error tends to increase as more fitting parameters are added, this upward trend is rarely monotonic; some parameter errors can decrease slightly as more fitting parameters are added, which could occur in part because of the stochasticity of the fitting process. Other parameters, especially ones with uniformly poor fits, have errors that are mostly independent of the number of fitting parameters, which could indicate weak correlation to the $I_d$-$V_{gs}$ profile.

**Supplementary Table S2: Median absolute errors for individual parameters when fitting HEMTs.** The number in the top row (5, 10, … 35) indicates the number of device parameters fit simultaneously, and the tabulated value indicates the median absolute error for each fitting parameter in a test set of 1,000 devices. All parameter names and units are identical to those in the ASM-HEMT model. A superscripted asterisk (*) after the parameter name denotes that the parameter was sampled in logarithmic space (as in **Supplementary Table S1**); for these parameters, the median absolute error was calculated by first taking the base-10 logarithm of the true and predicted values. Parameter errors are listed as hyphens (-) when that parameter is held constant, i.e., when the parameter was set to a fixed value, rather than varying it.

| **Number of fitting parameters** | **5** | **10** | **15** | **20** | **25** | **30** | **35** |
|---|---|---|---|---|---|---|---|
| **Parameter name** | | | | | | | |
| mexpaccs | 0.961 | 0.984 | 0.955 | 0.885 | 0.954 | 0.908 | 0.956 |
| thesat | $6.70\times10^{-4}$ | $2.81\times10^{-3}$ | $5.02\times10^{-2}$ | 0.136 | 0.153 | 0.305 | 0.345 |
| voff | $1.19\times10^{-4}$ | $4.40\times10^{-4}$ | $2.16\times10^{-3}$ | $1.89\times10^{-3}$ | $5.98\times10^{-3}$ | $7.63\times10^{-3}$ | $5.88\times10^{-3}$ |
| vsataccs | $3.31\times10^{4}$ | $3.59\times10^{4}$ | $3.45\times10^{4}$ | $3.13\times10^{4}$ | $3.21\times10^{4}$ | $2.98\times10^{4}$ | $3.01\times10^{4}$ |
| ub* | 0.532 | 0.773 | 0.880 | 0.993 | 1.18 | 1.13 | 1.18 |
| eta0 | - | $2.72\times10^{-4}$ | $5.78\times10^{-4}$ | $5.48\times10^{-4}$ | $6.98\times10^{-4}$ | $1.52\times10^{-3}$ | $1.13\times10^{-3}$ |
| igddio | - | 0.198 | 0.197 | 0.345 | 0.559 | 1.57 | 1.41 |
| njgd | - | 3.80 | 3.96 | 4.35 | 4.39 | 4.28 | 4.38 |
| rdc | - | $6.63\times10^{-6}$ | $2.65\times10^{-5}$ | $4.88\times10^{-5}$ | $1.21\times10^{-4}$ | $1.48\times10^{-4}$ | $1.71\times10^{-4}$ |
| rnjgd | - | $6.52\times10^{-2}$ | $5.85\times10^{-2}$ | 0.112 | 0.152 | $9.05\times10^{-2}$ | $7.05\times10^{-2}$ |
| nfactor | - | - | $5.33\times10^{-3}$ | $5.27\times10^{-3}$ | $1.11\times10^{-2}$ | $1.53\times10^{-2}$ | $1.14\times10^{-2}$ |
| njgs | - | - | 4.54 | 4.41 | 4.38 | 4.48 | 4.49 |
| u0accs | - | - | $1.30\times10^{-2}$ | $4.75\times10^{-2}$ | $5.01\times10^{-2}$ | $4.68\times10^{-2}$ | $4.84\times10^{-2}$ |
| ua | - | - | $4.52\times10^{-9}$ | $1.02\times10^{-8}$ | $2.99\times10^{-8}$ | $4.01\times10^{-8}$ | $4.00\times10^{-8}$ |
| imin* | - | - | 0.729 | 0.781 | 0.720 | 0.761 | 0.708 |
| delta | - | - | - | 0.236 | 0.498 | 0.597 | 0.568 |
| gdsmin* | - | - | - | 0.221 | 0.256 | 0.472 | 0.350 |
| igsdio | - | - | - | 1.86 | 1.77 | 1.89 | 1.97 |
| lambda | - | - | - | $4.46\times10^{-4}$ | $6.69\times10^{-4}$ | $7.48\times10^{-4}$ | $8.77\times10^{-4}$ |
| ns0accs* | - | - | - | 0.332 | 0.417 | 0.512 | 0.489 |
| mexpaccd | - | - | - | - | 0.954 | 0.913 | 0.873 |
| rigsdio | - | - | - | - | $2.30\times10^{-8}$ | $2.36\times10^{-8}$ | $2.27\times10^{-8}$ |
| u0 | - | - | - | - | $1.66\times10^{-2}$ | $2.11\times10^{-2}$ | $1.96\times10^{-2}$ |
| uted | - | - | - | - | 2.94 | 3.07 | 3.30 |
| ns0accd* | - | - | - | - | 0.779 | 0.831 | 0.676 |
| rigddio | - | - | - | - | - | $6.84\times10^{-9}$ | $6.47\times10^{-9}$ |
| rsc | - | - | - | - | - | $1.70\times10^{-4}$ | $1.80\times10^{-4}$ |
| rth0 | - | - | - | - | - | 6.35 | 6.49 |
| utes | - | - | - | - | - | 2.96 | 2.96 |
| vdscale | - | - | - | - | - | 0.114 | 0.135 |
| cdscd | - | - | - | - | - | - | $1.41\times10^{-3}$ |
| rnjgs | - | - | - | - | - | - | 2.47 |
| u0accd | - | - | - | - | - | - | $4.63\times10^{-2}$ |
| ute | - | - | - | - | - | - | 0.178 |
| vsat | - | - | - | - | - | - | $2.13\times10^{4}$ |

## Supplementary References